\documentclass{article}

    \PassOptionsToPackage{numbers, compress}{natbib}

 \usepackage[dblblindworkshop, final]{neurips_2026}
\workshoptitle{Personalized, Aligned, Long-Term Memory for AI Systems (PALM)}

\usepackage[utf8]{inputenc} 
\usepackage[T1]{fontenc}    
\usepackage{hyperref}       
\usepackage{url}            
\usepackage{booktabs}       
\usepackage{amsfonts}       
\usepackage{nicefrac}       
\usepackage{microtype}      
\usepackage{xcolor}         
\usepackage{amsmath}        
\usepackage{cleveref}
\usepackage{graphicx}
\graphicspath{{./}}
\usepackage{subcaption}
\usepackage{wrapfig}
\title{Continuous Memory Machines}

\author{%
  \textbf{Ciaran Regan\textsuperscript{1,2}}\thanks{Corresponding author: \href{mailto:ciaran@sakana.ai}{\texttt{ciaran@sakana.ai}}.}\quad
  \textbf{Kai Arulkumaran\textsuperscript{2}}\quad
  \textbf{Luke Darlow\textsuperscript{2}} \\
  \textbf{Stefania Druga\textsuperscript{2}}\quad
  \textbf{Sebastian Risi\textsuperscript{2}}\quad
  \textbf{Llion Jones\textsuperscript{2}} \\[0.5em]
  \textsuperscript{1}University of Tsukuba\quad
  \textsuperscript{2}Sakana AI
}

\begin{document}

\maketitle

\begin{abstract}
Recurrent neural networks typically compress information into a single vector-valued recurrent state, forcing short-term computation and long-term retention to share the same representation. Past extensions alleviate this bottleneck by increasing the memory capacity or separating timescales, but lack the combination of rapid neuron-level processing and longer-term retention found in biology. To that end, we introduce the Continuous Memory Machine (CMM), a recurrent architecture with matrix-valued short- and long-term memory states serving distinct functional roles. Building on the Continuous Thought Machine (CTM), the CMM's short-term memory tracks recent neural activity, with uniquely parameterized neuron-level models learning to use these activity patterns for computation. A persistent long-term memory stores information for later use, with a Transformer jointly updating both memory stores, providing an expressive bidirectional read--write mechanism such that each store can reorganize its own contents and both read from and write to the other. Across algorithmic, in-context learning, and recurrent reasoning tasks, the CMM outperforms a broad suite of baselines, exhibiting stronger generalization than prior memory-augmented networks while preserving the CTM's interpretable attention patterns. Code is available at \url{https://github.com/SakanaAI/continuous-memory-machines}.
\end{abstract}

\section{Introduction}

Many complex problems require sequential information to be maintained and manipulated over multiple timescales. A model executing an algorithm, for instance, needs to simultaneously track rapidly changing inputs while managing information acquired earlier. Such tasks can be challenging for Transformers~\citep{vaswani2017attention} which, despite their dominance in sequence processing, can struggle with state tracking~\citep{deletang2022neural}. Recurrent neural networks (RNNs), on the other hand, are naturally suited to such settings, updating a persistent hidden state over time. This state, however, typically compresses the entire sequence history into a single hidden vector~\citep{hochreiter1997long, chung2014empirical}, with the same representation responsible for both short-term computation and long-term memory. This creates a bottleneck that limits performance on tasks that may benefit from both rapid computation and persistent storage~\citep{graves2014neural}.

Several lines of work have sought to address these limitations. Multiscale RNNs introduce hidden states that update with different rates~\citep{schmidhuber1991neural, koutnik2014clockwork}, allowing some components to respond to recent inputs while others retain information over longer horizons. Fast weight RNNs take an alternative approach, replacing vector-valued with matrix-valued hidden states, supporting associative storage~\citep{schmidhuber1992fw,schlag2021linear,yang2024gated}. Memory-augmented neural networks (MANNs) couple a controller, such as a long short-term memory~\citep[LSTM;][]{hochreiter1997long}, to an addressable matrix, providing an external store from which information can be retrieved and updated~\citep{graves2014neural,weston2014memory}. Recurrent Transformers extend the effective context of standard Transformers by carrying hidden representations or memory tokens across successive input blocks~\citep{dai2019transformer,bulatov2022recurrent}. These approaches expand capacity or separate timescales, but the combination of persistent memory with rich short-term neural dynamics remains relatively unexplored.

Biological neural networks present this as a possible solution, with memory organized across interacting dynamical systems with distinct functional roles~\citep{mcclelland1995there,squire2004memory}. The complementary learning systems framework argues that this separation is necessary, as a single memory store cannot simultaneously encode new information and retain accumulated knowledge stably without interference~\citep{mcclelland1995there,kumaran2016learning}. Existing RNN architectures fail to capture this sufficiently. Multiscale RNNs retain vector-valued memories, fast weight RNNs increase the capacity of a single recurrent state, and MANNs pair an external matrix-valued memory with a limited controller state. Conversely, the recently-introduced Continuous Thought Machine~\citep[CTM;][]{darlow2025continuous} demonstrates the benefits of rich short-term dynamics, directly leveraging recent neural activation patterns for computation, but lacks long-term storage, as we show in~\cref{sec:experiments}. This motivates the question: ``Can the short-term dynamics of the CTM be combined with the capabilities of memory-augmented architectures to create a more capable recurrent model?''

To that end, we introduce the Continuous Memory Machine (CMM), a recurrent architecture that explicitly separates short- and long-term memory into distinct matrix-valued recurrent states. The CMM combines the persistent memory mechanisms of MANNs with the CTM's ability to exploit rich short-term neural dynamics for computation. Its short- and long-term memory states are processed by a shared Transformer, allowing each memory to update its own contents while also exchanging information with the other through learned bidirectional interactions. Across algorithmic, in-context learning, maze-solving, and image-classification tasks, the CMM achieves strong performance and improved generalization while preserving the CTM's interpretable reasoning behavior. Analysis reveals the CMM learns task-dependent memory strategies, relying on persistent long-term memory for memory-intensive tasks while suppressing cross-memory interaction when it is unnecessary. In summary, our contributions are as follows:

\begin{itemize}
\item We introduce the Continuous Memory Machine, an RNN with distinct matrix-valued short- and long-term memories. A shared Transformer updates both, supporting within-memory computation and learned cross-memory interaction.
\item We show the CMM achieves strong performance in algorithmic, few-shot learning and reasoning tasks, with improved generalization while preserving the interpretability of the CTM.
\item We show that the CMM learns task-dependent memory strategies, using long-term memory for retention and manipulation while suppressing interactions when persistent storage is unnecessary.
\end{itemize}

\begin{figure}[t!]
    \centering
    \includegraphics[width=\linewidth]{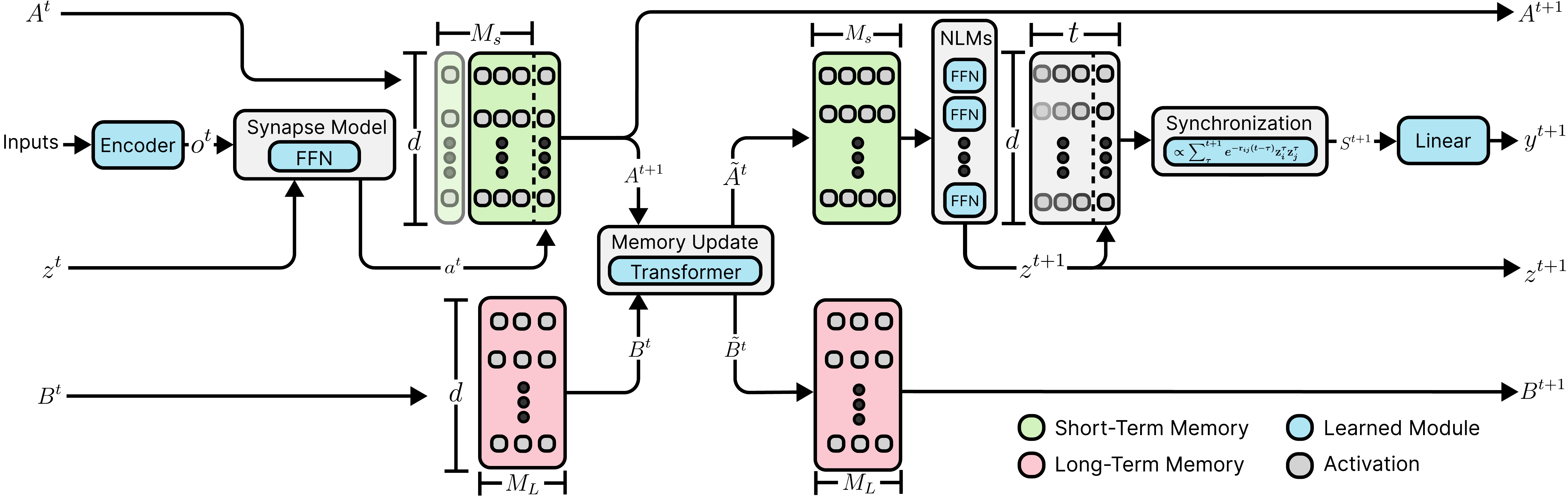}
    \caption{The Continuous Memory Machine. At timestep $t$, the current observation $o^t$ and previous activated state $z^t$ are processed by the feedforward \textit{synapse model} to produce a pre-activated state $a^t$. This state is appended to the sliding short-term memory (STM) $A^t$. $A^t$ and the long-term memory $B^t$ are jointly processed by a Transformer, producing updated memories $\tilde{A}^t$ and $\tilde{B}^{t}$. $\tilde{A}^t$ is passed through the \textit{neuron-level models} to produce the next activated state $z^{t+1}$. From the activated-state history, the model computes the synchronization $S^{t+1}$ and output $y^{t+1}$. $A^t$, $\tilde{B}^t=B^{t+1}$, and $z^{t+1}$ are carried forward to the next iteration.}

    \label{fig:architecture}
\end{figure}

\section{Related Work}

\subsection{Recurrent Neural Networks}
\textbf{Recurrent neural networks}~\citep[RNNs;][]{robinson1987utility,werbos1988generalization,williams1989complexity,elman1990finding} model temporal sequences by maintaining a hidden state that is updated over time. LSTMs introduced gating mechanisms to mitigate vanishing gradients, and, together with Gated Recurrent Units~\citep[GRUs;][]{chung2014empirical}, have become the standard recurrent architectures for sequence modeling. However, these models compress the sequence history into a fixed-size hidden vector, limiting how much information can be retained~\citep{hochreiter2001gradient}. Extensions address this limitation by modifying the structure of the hidden state or its update dynamics.

\textbf{Multiscale RNNs} aim to capture long-term dependencies through recurrent dynamics operating at different timescales~\citep{schmidhuber1991neural}. By updating less frequently, slower recurrent states can preserve information over longer horizons while shortening the effective gradient path between temporally distant inputs. Early work by~\citet{schmidhuber1991neural,schmidhuber1992learning} proposed a hierarchy of RNN layers running at different temporal resolutions.~\citet{koutnik2014clockwork} propose Clockwork RNNs, where the hidden layer is partitioned into separate modules, each processing inputs at a pre-specified temporal granularity. Follow-up work allowed multiscale RNNs to learn this granularity~\citep{chung2016hierarchical}. Fast-Slow RNNs split processing specifically into two interacting streams, with a slowly evolving higher-level recurrent module and faster lower-level modules, enabling long-term information to be retained while preserving sensitivity to recent inputs~\citep{mujika2017fast}.

\textbf{Fast weight RNNs} expand the recurrent state of standard RNNs from a vector to a matrix, increasing its capacity for associative storage~\citep{hinton1987using,schmidhuber1992fw, irie2025fast}. An early example is the Fast Weight Programmer~\citep{schmidhuber1992fw}, which writes key--value associations into a memory matrix and retrieves them using query vectors. This can be interpreted as linear attention~\citep{schlag2021linear}, which has motivated a range of modern recurrent architectures~\citep{schlag2021linear,gu2021efficiently,smith2022simplified,gu2023mamba}. xLSTM, for example, introduces an LSTM variant with matrix-valued memory~\citep{beck2024xlstm}, while the Gated Delta Network~\citep[GDN;][]{yang2024gated} combines gating with a delta-rule~\citep{widrow1994adaptive} update to selectively modify matrix-valued state.

\textbf{MANNs} augment RNNs with external matrix-valued memory stores~\citep{weston2014memory, sukhbaatar2015end, joulin2015inferring, santoro2016meta, wu2018kanerva}. Neural Turing Machines~\citep[NTMs;][]{graves2014neural} couple a neural controller with an addressable memory, analogous to the random-access memory coupled to a CPU in a computer, and access it through content- and location-based attention. Differentiable Neural Computers~\citep[DNCs;][]{graves2016hybrid} extend this with temporal and usage-based addressing, enabling the model to learn algorithms such as sorting through backpropagation. Although this improves retention, they can struggle with reasoning tasks as the stored memories do not interact with each other. To address this, \citet{santoro2018relational} introduced the Relational Memory Core (RMC), allowing memory slots to interact via self-attention. Building on the idea that attention itself can serve as a memory primitive, \citet{nam2023neural} reformulate attention as a readable and writable memory, using it to construct MANNs (LSAM and NAM-TM) that matched or exceeded the DNC's performance with simpler primitives.

\textbf{Recurrent Transformers} introduce memory mechanisms to the Transformer~\citep{vaswani2017attention} that carry information across input blocks~\cite{hutchins2022block, didolkar2022temporal}. Transformer-XL~\citep{dai2019transformer} caches hidden representations from previous blocks and makes them available to subsequent attention operations, extending the context. The Recurrent Memory Transformer~\citep[RMT;][]{bulatov2022recurrent} instead introduces dedicated memory tokens whose updated representations are passed between blocks, allowing the model to learn how information is retained and retrieved.

\subsection{Biological Memory}
\text{Biological memory} is not organized as a single store but instead as a collection of dynamical systems operating at different timescales~\citep{squire2004memory}. The complementary learning systems framework~\citep{mcclelland1995there,kumaran2016learning} argues that this organization is computationally necessary, since a single memory store cannot simultaneously encode new experiences quickly and retain accumulated knowledge stably without one interfering with the other. Brains avoid this trade-off by pairing a fast hippocampal system and a slow neocortical system~\citep{mcclelland1995there,kumaran2016learning}, with a similar split appearing within working memory itself~\citep{stokes2015activity,barbosa2020interplay}. Across both timescales, information is represented through coordinated neural dynamics~\citep{singer1999neuronal,fries2005mechanism}. While prior work in artificial neural networks has exploited neural dynamics for computation~\citep{miyato2024artificial,darlow2025continuous}, their integration across distinct memory timescales remains underexplored.

\section{Method}
We introduce the Continuous Memory Machine (CMM), a recurrent architecture with distinct matrix-valued short- and long-term hidden states with different functional roles. The CMM builds on the CTM, draws on the memory mechanisms of MANNs, and, similar to multiscale RNNs, separates timescales to support distinct retention strategies. A Transformer jointly updates the two stores, providing an expressive mechanism for retrieving, writing, and reorganizing memories, both within and across stores. An overview of the architecture is shown in~\cref{fig:architecture}. We review the CTM before introducing the CMM architecture.

\subsection{Continuous Thought Machines}
\label{sec:ctm}

The CTM is a recurrent architecture whose latent representation is grounded in neural dynamics. It combines two key ideas. First, \textit{neuron-level temporal processing}: each neuron independently processes a sliding window of its own activation history through a learned transformation, allowing individual neurons to discover distinct temporal features. Second, \textit{neural synchronization} as the latent representation: pairwise temporal correlations between neuron activities serve as the model's internal state, providing a smoothly evolving, high-dimensional representation. The CTM architecture is depicted in Appendix~\cref{fig:ctm_sideways_appendix}.

Concretely, the CTM processes information as follows. At each timestep $t$, the model receives the concatenation of its current activated state $z^t \in \mathbb{R}^{d}$ and an external observation $o^t \in \mathbb{R}^{d_\text{obs}}$, where $d$ and $d_{\text{obs}}$ denote the dimensionalities of the hidden and input vectors, respectively. A feedforward \textit{synapse} network produces a pre-activated state $a^t$:
\begin{equation}
    a^t = f(z^t, o^t) \in \mathbb{R}^d,
    \label{eq:synapse}
\end{equation}
where $f$ is a learned feedforward network. This pre-activated state is appended to a sliding window of the $M_S$ most recent pre-activations, forming a matrix $A^{t} \in \mathbb{R}^{d \times M_S}$. In this work, we refer to this matrix as the \textit{Short-Term Memory} (STM) of the model. Each neuron $i$ then independently maps its row of this history to the next activated state via a \textit{neuron-level model} (NLM):
\begin{equation}
    z_i^{t+1} = g_i(A_i^t),
    \label{eq:nlm}
\end{equation}
where $g_{i}$ is a small feedforward network with parameters unique to each neuron $i$. These NLMs give each neuron a learned temporal response based on its historical activity, which was shown to produce more diverse activity patterns and lead to benefits in reasoning tasks~\citep{darlow2025continuous}. The CTM's latent representation is \textit{synchronization}, computed as an exponentially weighted dot product of activated state histories:
\begin{equation}
    S^{t+1}_{ij} \;\propto\;
    \sum_{\tau=1}^{t+1} \mathrm{e}^{-r_{ij}(t+1-\tau)}\, z^{\tau}_i\, z^{\tau}_j,
    \label{eq:synch}
\end{equation}

where $r_{ij}$ is a learned decay parameter for each neuron pair, controlling the influence of past states. The output at each timestep is a mapping from the synchronization matrix, $y^{t+1} = W_{\text{out}}\text{vec}(S^{t+1})$. Since $A^t$ retains only a finite window of recent activity and $S^t$ compresses the activation history into a decaying summary, the CTM has no persistent mechanism for storing information over long horizons and, as we show empirically in~\cref{exp:algo}, this substantially limits its performance on tasks that require long-term retention.

\subsection{Continuous Memory Machines}
\label{sec:cmm}
Building upon the CTM, we introduce an explicit matrix-valued long-term memory that operates alongside its sliding-window STM. The two memories are jointly updated by a Transformer, providing an expressive mechanism for writing, retrieval, and reorganization across stores.

\textbf{Long-Term Memory.} We define the \textit{Long-Term Memory} (LTM) as a matrix $B^t \in \mathbb{R}^{d \times M_L}$, whose $M_L$ columns are treated as tokens of dimension $d$. The LTM is initialized with a learned matrix $B^0$ and is subsequently updated at every timestep through interaction with the STM.

\textbf{Joint Memory Update.} At each timestep, the CMM first updates the STM as in the CTM. The synapse produces a pre-activated state $a^t$ from the previous activated state $z^t$ and the current observation $o^t$ (\cref{eq:synapse}), which is appended to the sliding window to form $A^t \in \mathbb{R}^{d \times M_S}$. Treating the columns of $A^t$ and $B^{t-1}$ as tokens of dimension $d$, we concatenate a learnable sink token $s \in \mathbb{R}^{d}$~\citep{xiao2023efficient}, the LTM, and the STM into a single sequence,
\begin{equation}
    X^t = \big[\,s\, ;\, B^{t-1}\, ;\, A^t\,\big]
    \in \mathbb{R}^{d \times (1 + M_L + M_S)},
    \label{eq:cmm_concat}
\end{equation}
add learned positional embeddings, and pass $X^t$ through a stack of $J$ Transformer blocks to obtain
\begin{equation}
    \tilde{X}^t
    = \mathrm{TransformerBlocks}(X^t)
    = \big[\,\tilde{s}^t\,;\,\tilde{B}^t\,;\,\tilde{A}^t\,\big],
    \label{eq:cmm_transformer}
\end{equation}
where $\tilde{B}^t \in \mathbb{R}^{d \times M_L}$ and $\tilde{A}^t \in \mathbb{R}^{d \times M_S}$ denote the updated long- and short-term memories.

\textbf{Activated State and Recurrence.} The updated STM $\tilde{A}^t$ is passed through the neuron-level models to produce the next activated state, as in the CTM (\cref{eq:nlm}):
\begin{equation}
    z^{t+1}_i = g_i\big(\tilde{A}^{t}_i\big).
    \label{eq:cmm_nlm}
\end{equation}
The neuron-level models therefore operate on the STM after its interaction with the LTM, conditioning each neuron's update on both its recent pre-activation history and information retrieved from long-term memory. The synchronization $S^{t+1}$ and output $y^{t+1}$ are then computed from the activated state histories as in \cref{eq:synch}.

To advance to the next timestep, the updated LTM $\tilde{B}^t$ is carried forward as the persistent state, while the updated STM $\tilde{A}^t$ is discarded after producing $z^{t+1}$. The original STM $A^t$ is retained as the rolling window of recent pre-activations and updated with the next synapse output.

\textbf{Relation to Prior Architectures.} The CMM differs from prior MANNs in how memory is accessed and organized. Whereas the DNC~\citep{graves2016hybrid} uses structured addressing and allocation mechanisms, the CMM learns memory interactions through self-attention, following the RMC~\citep{santoro2018relational}. Unlike the RMC, the CMM maintains distinct short- and long-term matrix-valued stores with different functional roles. The LTM retains and reorganizes information across timesteps, while the NLMs extract temporal features from the transformed STM histories to compute the next state. Their private parameters allow each neuron to learn a distinct response to recent activity, which has been shown to aid reasoning~\citep{darlow2025continuous}. Recurrent Transformers such as RMT provide persistent memory but lack this explicit neuron-level temporal processing. Finally, the sink token enables flexible memory use by allowing the model to suppress attention to either store. As we show in~\cref{sec:analysis}, the CMM learns different memory strategies across tasks, relying on persistent LTM for memory-intensive tasks while bypassing LTM when not needed.

\section{Experiments}
\label{sec:experiments}
Our experiments address two questions. First, does the CMM improve memory manipulation and generalization over existing architectures? Second, does it improve the CTM's reasoning while preserving its interpretable dynamics? We address these questions through algorithmic, maze-solving, in-context learning and image-classification tasks. Unless otherwise stated, we report the mean and standard deviation across three random seeds.

\paragraph{Baselines}\label{sec:baselines}
We compare the CMM against ten baselines chosen to isolate distinct design choices. To isolate the contribution of the long-term memory mechanism, we include the original CTM~\citep{darlow2025continuous}. As standard RNN baselines we include a ``vanilla'' RNN and LSTM~\citep{hochreiter1997long} with layer normalization~\citep[LayerNorm;][]{ba2016layer}. For a multiscale RNN baseline we use the Fast-Slow LSTM~\citep[FS-LSTM;][]{mujika2017fast}. For a modern matrix-valued hidden state RNN we use the GDN~\citep{yang2024gated}, allowing for negative eigenvalues for improved state tracking capabilities~\citep{grazzi2025unlocking}. For MANNs, we include the DNC~\citep{graves2016hybrid} (superseding the NTM~\citep{graves2014neural}), the RMC~\citep{santoro2018relational}, and the more recent LSAM and NAM-TM~\citep{nam2023neural}. For a recurrent Transformer we include the RMT~\citep{bulatov2022recurrent}. In addition, we report a causal Transformer~\cite{vaswani2017attention} as a non-recurrent reference. Unlike the bounded-memory recurrent baselines, it directly attends over the full context, giving it a strong advantage.

\paragraph{Ablations} We additionally compare against three parameter-matched architectural ablations. CMM-LSTM replaces the CTM backbone with a LayerNorm LSTM while retaining the LTM and memory-update Transformer, testing the contribution of the CTM's short-term dynamics. CTM-Ext.\ extends the STM window to match the CMM's total memory slots. CMM-STM is a CMM with $M_L=0$, testing the CTM's STM combined with a Transformer, without persistent LTM. Comparing CTM-Ext.\ and CMM-STM tests the benefit of the Transformer on the STM. Comparing CMM-STM and CMM tests the benefit of a distinct persistent LTM. The Appendix contains detailed architecture comparisons (\cref{app:arch-comparison}), sweeps of LTM size $M_L$ and Transformer depth $J$ (\cref{app:sweeps}) and lists all hyperparameters.

\subsection{Memory Manipulation: Algorithmic Tasks}
\label{exp:algo}
\begin{figure}[!t]
    \centering
    \includegraphics[width=1.0\linewidth]{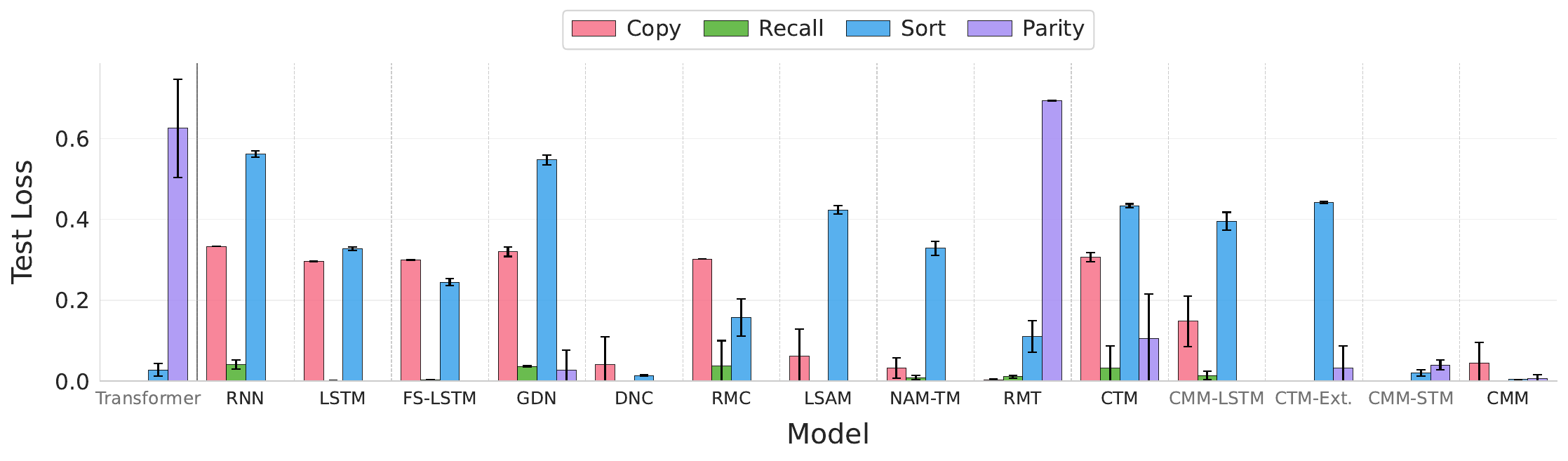}
    \caption{Loss for algorithmic tasks. The CMM demonstrates strong performance on all four tasks.}
    \label{fig:algorithmic-tasks-a}
\end{figure}

We evaluate on algorithmic tasks which have been problematic for standard RNNs~\citep{graves2014neural}. Test losses are reported in \cref{fig:algorithmic-tasks-a}.

\textbf{Copy.} The copy task tests whether a model can store and recall a long sequence of arbitrary information~\citep{graves2014neural}. The model receives $100$ vectors of dimension $D=10$ with entries drawn uniformly from $[-1, 1]$ and must reproduce the sequence from memory. The loss is the mean squared error computed over the output values. The non-recurrent Transformer attains a near-zero loss, as expected given its direct access to the complete sequence. Among recurrent baselines, the CMM, DNC, and NAM-TM all achieve losses below $5{\times}10^{-2}$, with LSAM close behind at $6.2{\times}10^{-2}$. The CTM, on the other hand, obtains $3.1{\times}10^{-1}$. This confirms that the CTM's short activation window is insufficient for sequence-level recall, and that adding the LTM closes the gap to architectures explicitly designed for algorithmic manipulation.

\begin{wrapfigure}{r}{0.45\textwidth}
    \vspace{-6pt}
    \centering
    \includegraphics[width=1\linewidth]{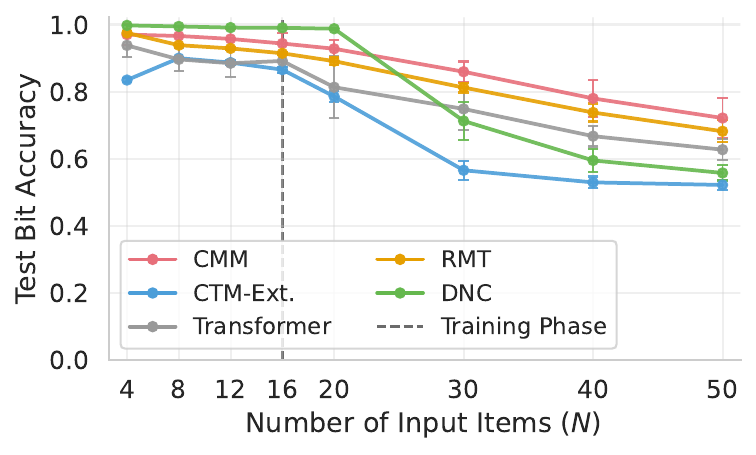}
    \caption{The CMM generalizes best on sort.}
    \label{fig:algorithmic-tasks-b}
\end{wrapfigure}

\textbf{Recall.} While the copy task probes sequential storage and retrieval, recall tests a more complex form of memory access in which one data item points to another~\citep{graves2014neural}. The model is shown a sequence of vectors, followed by a query equal to one of them, and must return the item that followed the query in the original sequence. The loss is the binary cross-entropy over the predicted item. The CMM, DNC, and LSAM all achieve test losses below $10^{-3}$, with the CMM reaching $99.95\%$ accuracy. The non-recurrent Transformer similarly obtains a loss of $2.7{\times}10^{-4}$ by directly attending over the complete sequence. By contrast, the CTM obtains a loss of $3.2{\times}10^{-2}$, over 60 times that of the CMM, reflecting the limits of its short activation window for retrieval.

\textbf{Sort.} This task tests whether the model can sort data, an important elementary algorithm~\citep{graves2014neural}. The model receives $N=20$ input items, each consisting of a scalar priority drawn uniformly from $[-1,1]$ and an associated $W(=8)$-bit binary value vector. After the input phase, the model must output the $K=16$ value vectors corresponding to the $K$ lowest-priority items, sorted in ascending order of priority. The loss is the binary cross-entropy computed over the $K$ output positions. The CMM attains the lowest test loss of all evaluated models at $4.1{\times}10^{-3}$, improving over the DNC at $1.4{\times}10^{-2}$ and the Transformer at $2.7{\times}10^{-2}$, while reducing the CTM's loss by two orders of magnitude. This is notable given that sort requires both retrieval and reorganization.

To test length generalization, we train on input lengths $N\in\{4,\ldots,16\}$ with $W=4$ and $K=4$, then evaluate up to $N=50$. The CMM, CTM-Ext., and DNC each use 26 memory elements, with $M_S=4$ and $M_L=22$ in the CMM, $M_S=26$ in CTM-Ext., and 26 external memory slots in the DNC. Results are shown in \cref{fig:algorithmic-tasks-b}. At the longest out-of-distribution length, the CMM achieves $72.1\%$ accuracy, compared with the RMT's $68.2\%$ and the DNC's near-chance $55.8\%$.

\textbf{Parity.} The model receives a sequence of random bits followed by a query token and predicts if the sequence is even or odd. We train models on sequence lengths 1--40 and test at length 40. All evaluated architectures except the Transformer and the RMT perform strongly on this task (\cref{fig:algorithmic-tasks-a}), due to the Transformer's limited state-tracking capabilities~\citep{deletang2022neural, hahn2020theoretical}.

Across copy, recall, and sort, CMM outperforms CMM-LSTM, showing that the CTM backbone adds value beyond the shared LTM and memory-update Transformer. CTM-Ext.\ and CMM-STM perform well on copy and recall because their extended STM keeps relevant inputs directly accessible. On sort, which requires selective retrieval and reordering, no ablation surpasses CMM. CMM-STM improves over CTM-Ext. due to the additional Transformer, but it still trails CMM, supporting the benefit of jointly using persistent LTM and transient STM.

\begin{figure}[t!]
  \centering
  \captionsetup{aboveskip=3pt,belowskip=3pt}

  \begin{minipage}[t]{0.55\linewidth}
    \vspace{0pt}
    \centering
    \captionof{table}{Maze-solving test accuracy.}
    \label{table:maze-results}
    \begin{small}
      \begin{sc}
        \setlength{\tabcolsep}{2pt}
        \begin{tabular}{@{}lcc@{}}
          \toprule
          Architecture & Action Acc. (\%) & Maze Acc. (\%) \\
          \midrule
          \textcolor{black!60}{Transformer} & \textcolor{black!60}{$41.4 \scalebox{0.85}{$\pm 6.9$}$} & \textcolor{black!60}{$0.7 \scalebox{0.85}{$\pm 0.5$}$} \\
          \midrule
          RNN & $47.2 \scalebox{0.85}{$\pm 2.2$}$ & $2.1 \scalebox{0.85}{$\pm 0.3$}$ \\
          LSTM & $50.6 \scalebox{0.85}{$\pm 10.1$}$ & $2.7 \scalebox{0.85}{$\pm 1.7$}$ \\
          FS-LSTM & $54.7 \scalebox{0.85}{$\pm 9.8$}$ & $3.3 \scalebox{0.85}{$\pm 1.9$}$ \\
          DNC & $82.7 \scalebox{0.85}{$\pm 2.6$}$ & $13.8 \scalebox{0.85}{$\pm 2.5$}$ \\
          RMC & $23.7 \scalebox{0.85}{$\pm 2.7$}$ & $0.0 \scalebox{0.85}{$\pm 0.0$}$ \\
          NAM-TM & $82.5 \scalebox{0.85}{$\pm 0.7$}$ & $13.2 \scalebox{0.85}{$\pm 0.5$}$ \\
          CTM & $88.8 \scalebox{0.85}{$\pm 8.3$}$ & $37.4 \scalebox{0.85}{$\pm 38.7$}$ \\
          \midrule
          \textcolor{black!60}{Ab: CMM-LSTM} & \textcolor{black!60}{$46.3 \scalebox{0.85}{$\pm 7.9$}$} & \textcolor{black!60}{$2.1 \scalebox{0.85}{$\pm 1.2$}$} \\
          \textcolor{black!60}{Ab: CMM-STM} & \textcolor{black!60}{$87.0 \scalebox{0.85}{$\pm 17.6$}$} & \textcolor{black!60}{$43.8 \scalebox{0.85}{$\pm 42.5$}$} \\
          CMM (Ours) & $\mathbf{96.3} \scalebox{0.85}{$\pm 3.8$}$ & $\mathbf{61.6} \scalebox{0.85}{$\pm 35.0$}$ \\
          \bottomrule
        \end{tabular}
      \end{sc}
    \end{small}
  \end{minipage}
  \hfill
 \begin{minipage}[t]{0.44\linewidth}
  \vspace{4pt}
  \centering

  \includegraphics[width=0.7\linewidth]
    {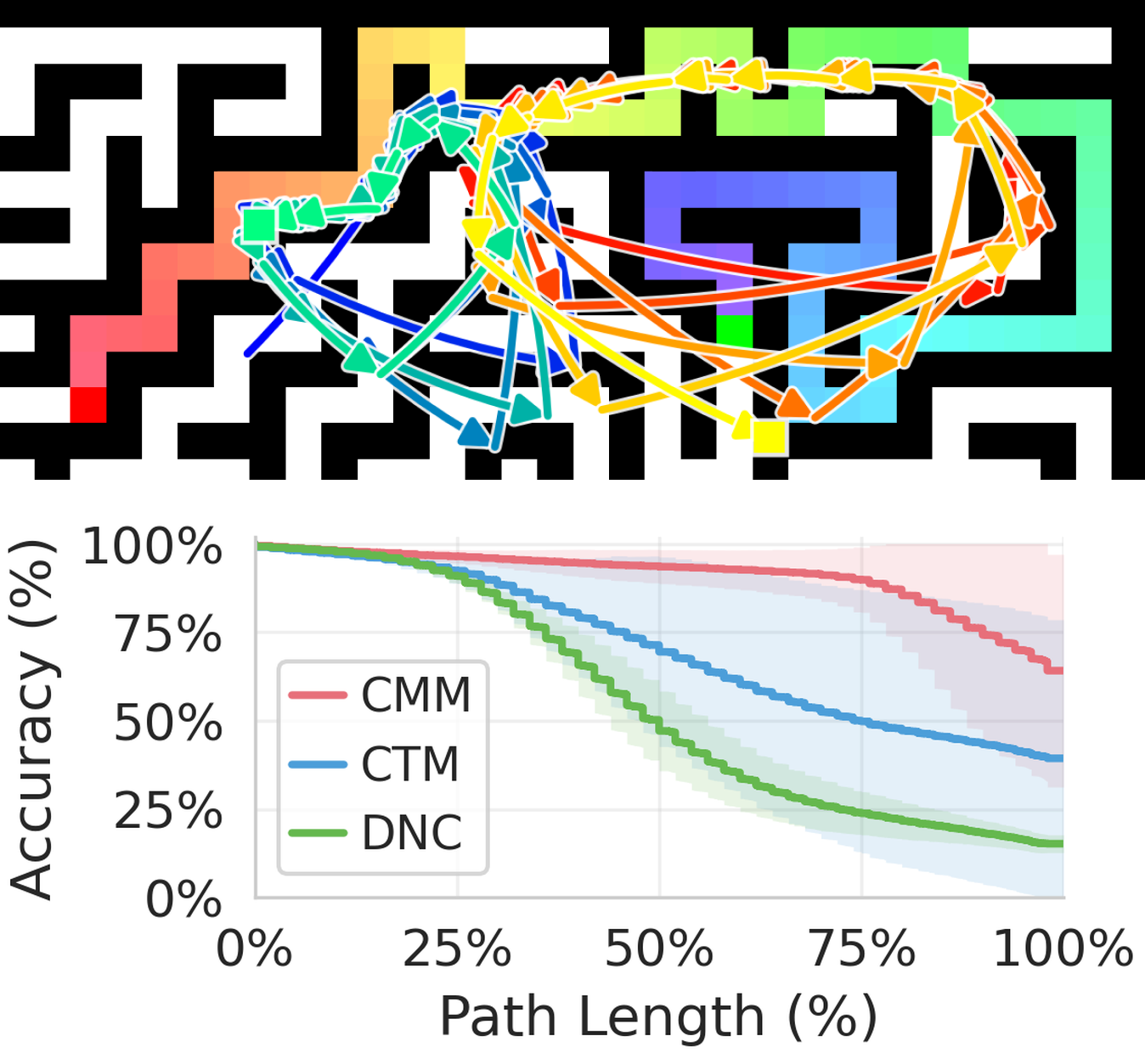}

  \captionof{figure}{
    Top: CMM attention traces a path through the $39\times39$ maze (crop).
    Bottom: Accuracy by maze path length. The CMM performs best on longer paths.
  }
  \label{fig:maze-combined}
\end{minipage}
\end{figure}

\subsection{Recurrent Reasoning: Maze-Solving}
We revisit the maze-solving task of~\citet{darlow2025continuous} to test whether the CMM improves the CTM’s recurrent reasoning capabilities while preserving its interpretable attention patterns. The model is given a $39\times39$ maze image with start and goal locations and must predict the sequence of actions connecting them. The model processes the maze over 50 internal steps, producing a full route prediction at each iteration. Following \citet{darlow2025continuous}, we omit positional embeddings from the maze image features, encouraging the model to learn an internal representation of the maze’s spatial structure. Experiments use a budget of 10M parameters, with full details provided in~\cref{app:maze}. 

We report action and full maze accuracies of held-out mazes in~\cref{table:maze-results}. The CMM outperforms all baselines\footnote{GDN, RMT, and LSAM could not train stably and are therefore excluded. See~\cref{app:maze}.}, including the CTM, increasing action accuracy from $88.8\%$ to $96.3\%$ and full maze accuracy from $37.4\%$ to $61.6\%$.  Although variability increases at the longest path lengths, and hence the greater uncertainty the full maze accuracy, the CMM performs markedly better at longer path lengths (\cref{fig:maze-combined}, bottom), with the strongest single run achieving $90.7\%$ full maze accuracy, with per-seed results are provided in~\cref{table:maze-results-per-seed}. Its gains over the ablations further highlight the benefit of combining the CTM's temporal processing with an expressive memory Transformer. Furthermore, the CMM preserves the CTM's path-following attention behavior (\cref{fig:maze-combined}, top), with the CMM's cross attention scanning a path through the maze while predicting the correct sequence of actions.

\newpage

\subsection{In-Context Learning: Few-Shot Regression}

\begin{wraptable}{r}{0.52\textwidth}
  \centering
  \caption{Few-shot regression accuracy with $K{=}10,20$ support examples. CMM achieves the highest accuracy in both settings.}
  \label{table:fewshot-regression-results}
  \begin{small}
    \begin{sc}
      \setlength{\tabcolsep}{3pt}
      \begin{tabular}{@{}lcc@{}}
        \toprule
        Architecture & $K{=}20$ & $K{=}10$ \\
        \midrule
        \textcolor{black!60}{Transformer} & \textcolor{black!60}{$29.75 \scalebox{0.85}{$\pm 1.24$}$} & \textcolor{black!60}{$9.09 \scalebox{0.85}{$\pm 0.08$}$} \\
        \midrule
        RNN & $5.39 \scalebox{0.85}{$\pm 0.13$}$ & $4.81 \scalebox{0.85}{$\pm 0.06$}$ \\
        LSTM & $9.72 \scalebox{0.85}{$\pm 0.42$}$ & $6.40 \scalebox{0.85}{$\pm 0.05$}$ \\
        FS-LSTM & $11.02 \scalebox{0.85}{$\pm 0.10$}$ & $6.53 \scalebox{0.85}{$\pm 0.07$}$ \\
        GDN & $10.65 \scalebox{0.85}{$\pm 0.94$}$ & $6.68 \scalebox{0.85}{$\pm 0.06$}$ \\
        DNC & $9.59 \scalebox{0.85}{$\pm 0.29$}$ & $6.34 \scalebox{0.85}{$\pm 0.06$}$ \\
        RMC & $19.53 \scalebox{0.85}{$\pm 7.93$}$ & $7.94 \scalebox{0.85}{$\pm 0.30$}$ \\
        LSAM & $29.82 \scalebox{0.85}{$\pm 0.18$}$ & $8.99 \scalebox{0.85}{$\pm 0.08$}$ \\
        NAM-TM & $27.65 \scalebox{0.85}{$\pm 0.11$}$ & $9.34 \scalebox{0.85}{$\pm 0.10$}$ \\
        RMT & $30.46 \scalebox{0.85}{$\pm 0.41$}$ & $9.81 \scalebox{0.85}{$\pm 0.08$}$ \\
        CTM & $8.78 \scalebox{0.85}{$\pm 0.11$}$ & $4.85 \scalebox{0.85}{$\pm 0.07$}$ \\
        \midrule
        \textcolor{black!60}{Ab: CMM-LSTM} & \textcolor{black!60}{$19.95 \scalebox{0.85}{$\pm 8.75$}$} & \textcolor{black!60}{$6.38 \scalebox{0.85}{$\pm 0.05$}$} \\
        \textcolor{black!60}{Ab: CTM-Ext.} & \textcolor{black!60}{$8.00 \scalebox{0.85}{$\pm 0.07$}$} & \textcolor{black!60}{$5.84 \scalebox{0.85}{$\pm 0.02$}$} \\
        \textcolor{black!60}{Ab: CMM-STM} & \textcolor{black!60}{$13.03 \scalebox{0.85}{$\pm 0.61$}$} & \textcolor{black!60}{$7.28 \scalebox{0.85}{$\pm 0.12$}$} \\
        CMM (Ours) & $\mathbf{30.63} \scalebox{0.85}{$\pm 0.24$}$ & $\mathbf{9.86} \scalebox{0.85}{$\pm 0.07$}$ \\
        \bottomrule
      \end{tabular}
    \end{sc}
  \end{small}
\end{wraptable}

\label{exp:fewshot-regression}
Few-shot learning requires a model to incorporate new examples while retaining previous information, making it a natural setting for evaluating short- and long-term memory~\citep{santoro2016meta}. We evaluate this capability using a few-shot regression task commonly used in meta-learning~\citep{finn2017model, duan2023hebbian}. In each episode a new target mapping $f:[-1,1]^{n}\to\mathbb{R}$ is sampled from a family of linear functions. The model is presented with $K$ noisy support pairs $(\mathbf{x}_t, f(\mathbf{x}_t) + \epsilon_t)$ with $\epsilon_t \sim \mathcal{N}(0,\sigma^2)$, followed by query inputs whose outputs it must predict. Since the target function changes between episodes, the model cannot memorize any single mapping during training and must instead learn an in-context inference procedure. We set the input dimension to $n=12$ and report results for $K\in\{10,20\}$. Full details are given in \cref{app:fewshot-regression}.

Test query accuracies across both support-set sizes are shown in~\cref{table:fewshot-regression-results}. A query prediction is correct when its absolute error is below $0.05$ on the standardized target scale. The CMM achieves the highest mean accuracy in both settings. At $K{=}20$, it improves accuracy over the CTM from $8.78\%$ to $30.63\%$ on linear functions. The CMM also achieves higher mean accuracies than all three architectural ablations, supporting the benefit of combining the CTM's sliding-window STM with a distinct persistent LTM.

\subsection{Image Classification: ImageNet-100}
\begin{figure}[t!]
  \centering
  \captionsetup{font=small,aboveskip=3pt,belowskip=3pt,justification=raggedright,singlelinecheck=false}

  \begin{minipage}[t]{0.55\linewidth}
    \vspace{0pt}
    \centering

    \captionof{table}{ImageNet-100 accuracy. The CMM is among the top performing baselines.}
    \label{table:imagenet-results}

    \begin{small}
      \begin{sc}
        \setlength{\tabcolsep}{3pt}
        \begin{tabular}{@{}lc@{}}
        \toprule
        Architecture & Acc. (\%) \\
        \midrule
          \textcolor{black!60}{Transformer}
            & \textcolor{black!60}{$89.5 \scalebox{0.85}{$\pm 0.2$}$} \\
          \midrule
          RNN
            & $89.6 \scalebox{0.85}{$\pm 0.2$}$ \\
          LSTM
            & $\mathbf{89.7 \scalebox{0.85}{$\pm 0.1$}}$ \\
          FS-LSTM
            & $89.5 \scalebox{0.85}{$\pm 0.1$}$ \\
          GDN
            & $63.3 \scalebox{0.85}{$\pm 41.3$}$ \\
          DNC
            & $\mathbf{89.7 \scalebox{0.85}{$\pm 0.2$}}$ \\
          RMC
            & $75.9 \scalebox{0.85}{$\pm 9.0$}$ \\
          LSAM
            & $59.2 \scalebox{0.85}{$\pm 49.0$}$ \\
          NAM-TM
            & $\mathbf{89.7 \scalebox{0.85}{$\pm 0.1$}}$ \\
          RMT
            & $89.5 \scalebox{0.85}{$\pm 0.1$}$ \\
          CTM
            & $89.4 \scalebox{0.85}{$\pm 0.3$}$ \\
          \midrule
          \textcolor{black!60}{Ab: CMM-LSTM}
            & \textcolor{black!60}{$89.3 \scalebox{0.85}{$\pm 0.1$}$} \\
          \textcolor{black!60}{Ab: CTM-Ext.}
            & \textcolor{black!60}{$\mathbf{89.7 \scalebox{0.85}{$\pm 0.0$}}$} \\
          \textcolor{black!60}{Ab: CMM-STM}
            & \textcolor{black!60}{$89.6 \scalebox{0.85}{$\pm 0.1$}$} \\
          CMM (Ours)
            & $\mathbf{89.7 \scalebox{0.85}{$\pm 0.2$}}$ \\
          \bottomrule
        \end{tabular}
      \end{sc}
    \end{small}
  \end{minipage}%
  \hfill%
  \begin{minipage}[t]{0.4\linewidth}
    \centering
    \setlength{\parskip}{0pt}
    \vspace{4pt}

    \includegraphics[width=0.49\linewidth,trim=7.2bp 7.2bp 7.2bp 7.2bp,clip]
      {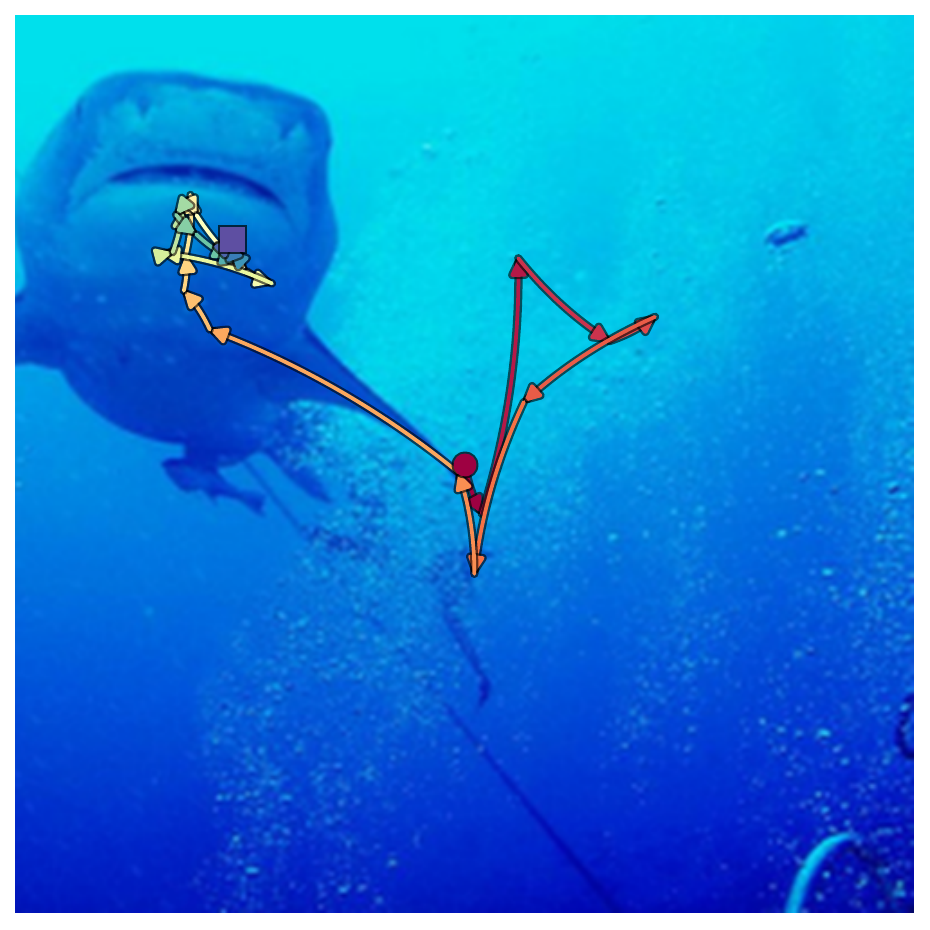}%
    \hfill%
    \includegraphics[width=0.49\linewidth,trim=7.2bp 7.2bp 7.2bp 7.2bp,clip]
      {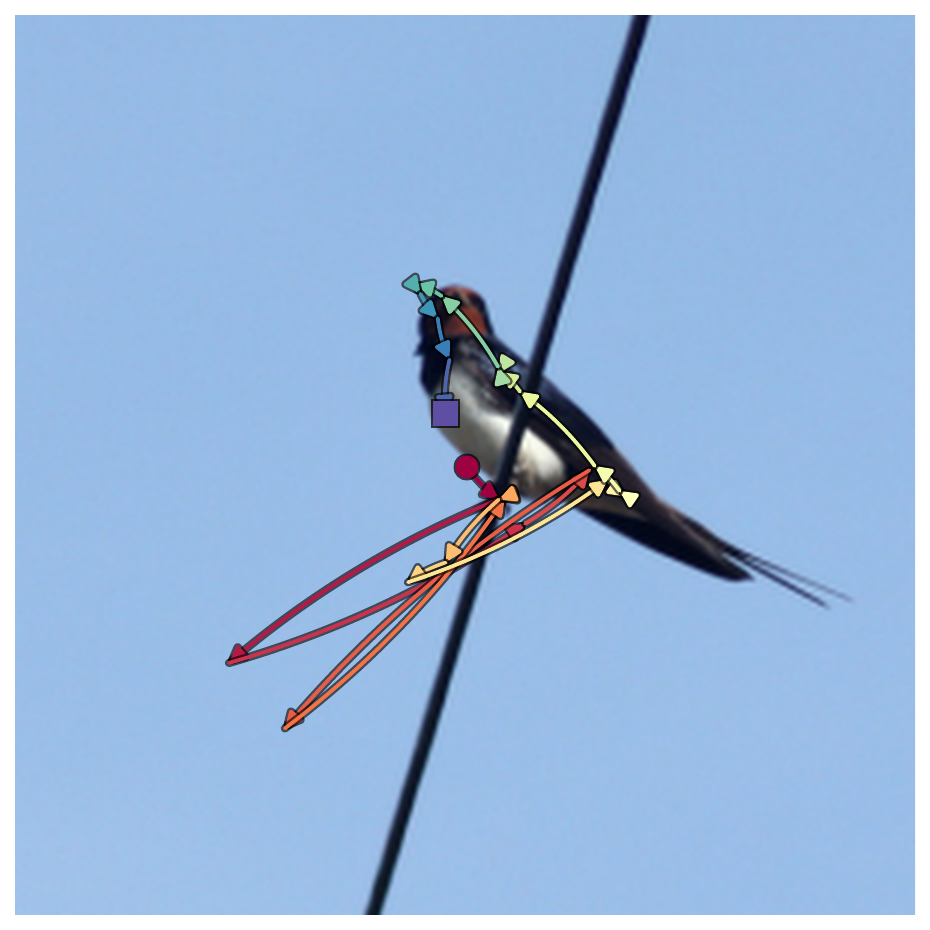}\par

    \nointerlineskip\vspace{4pt}
    \includegraphics[width=0.49\linewidth,trim=7.2bp 7.2bp 7.2bp 7.2bp,clip]
      {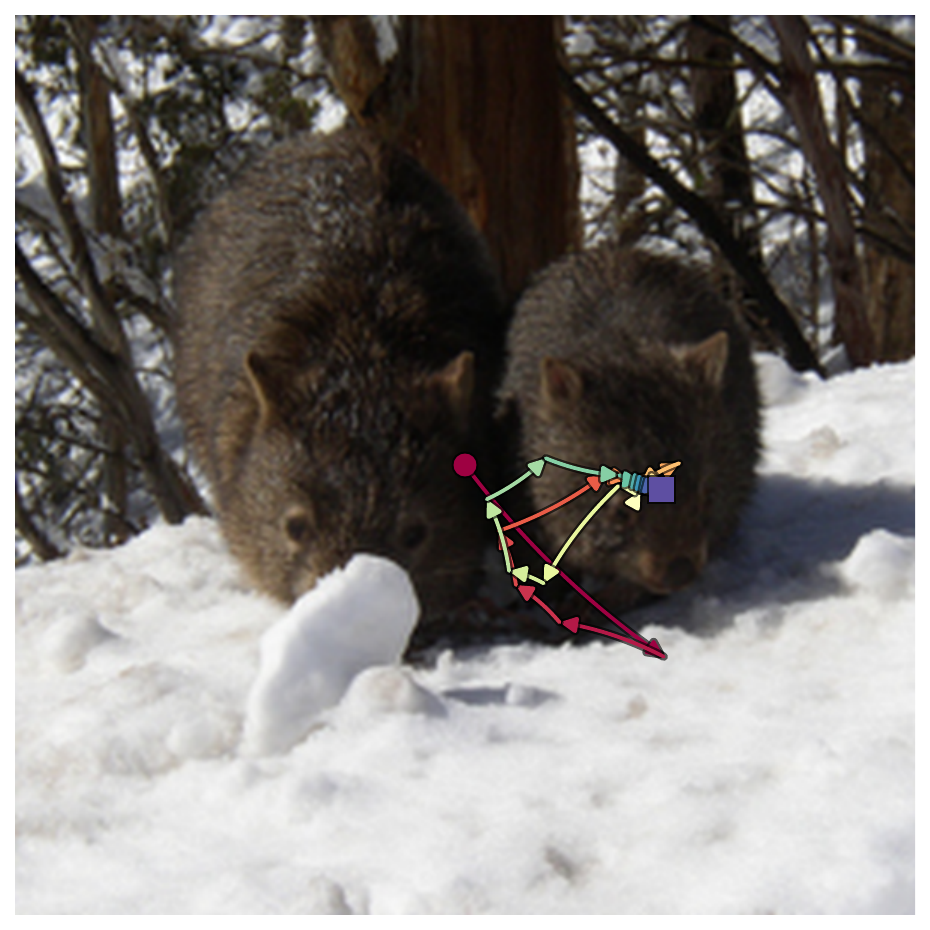}%
    \hfill%
    \includegraphics[width=0.49\linewidth,trim=7.2bp 7.2bp 7.2bp 7.2bp,clip]
      {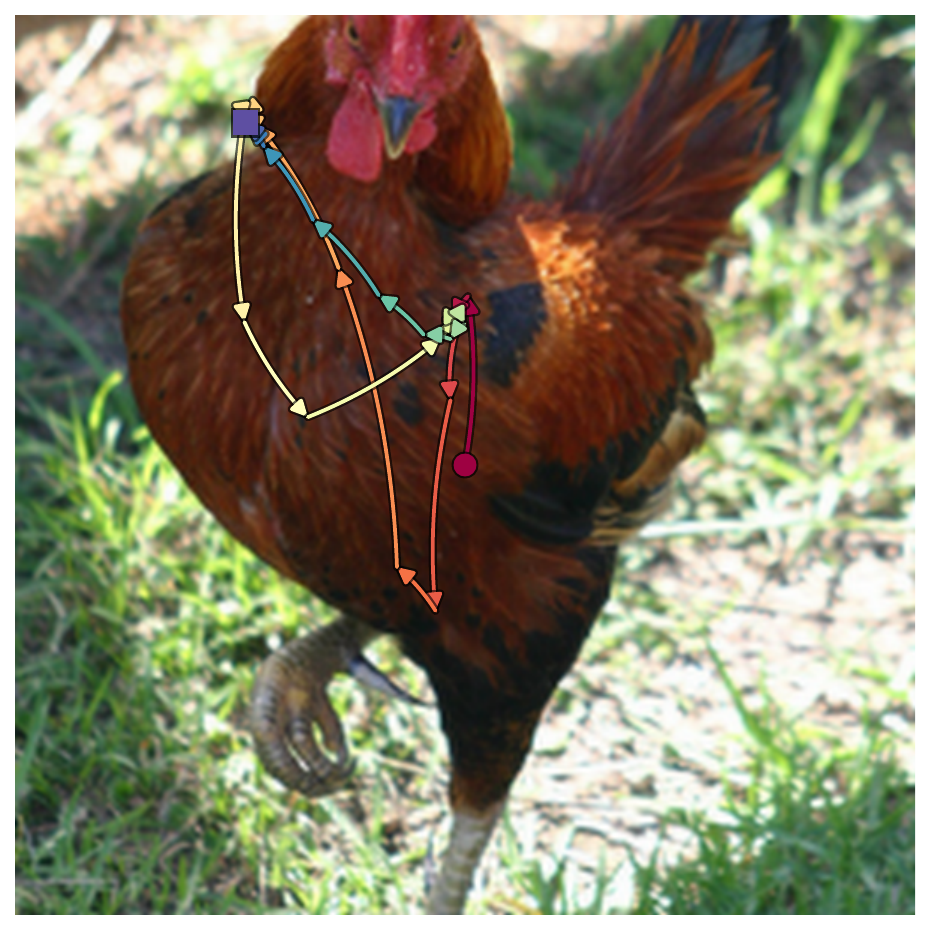}\par

    \captionof{figure}{CMM attention exhibits the CTM's ``looking around'' behavior \citep{darlow2025continuous}.}
    \label{fig:imagenet}
    \label{fig:imagenet-attn}
  \end{minipage}
\end{figure}

Following~\citet{darlow2025continuous}, we use image classification to examine how models gather visual information over successive computation steps, rather than to pursue state-of-the-art classification accuracy. We trained models with 1M-parameter recurrent heads using the first two blocks of a frozen pre-trained ViT as the backbone~\citep{wang2026vit}, with patch features serving as keys and values for cross-attention from synchronization-based queries. Results are shown in~\cref{table:imagenet-results} and~\cref{fig:imagenet-attn}. We find that the CMM is among the best performing architectures, while exhibiting interpretable attention patterns that shift between regions. Experimental details are provided in~\cref{app:imagenet}.

\subsection{Analysis: Task-Dependent Memory Usage}
\label{sec:analysis}
\begin{figure}[htbp!]
  \centering
  \includegraphics[width=\linewidth]{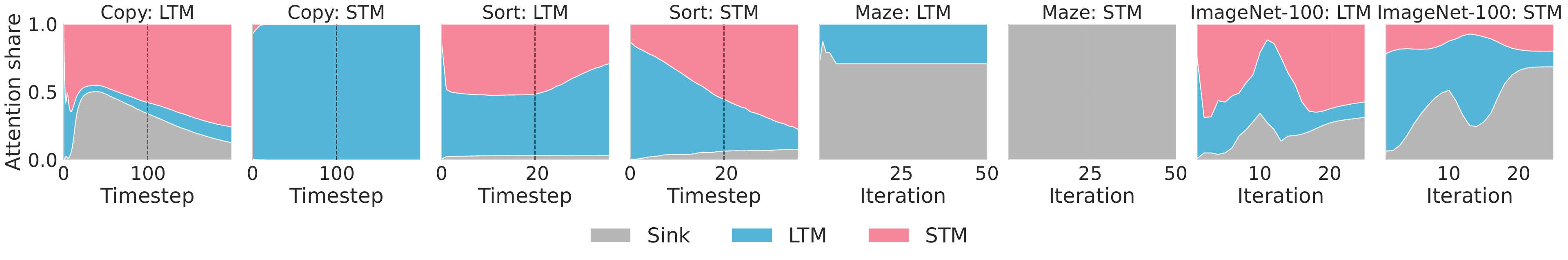}
  \caption{Memory attention across tasks. Each pair shows LTM queries (left) and STM queries (right), with attention to the sink, LTM, and STM tokens. CMM learns task-specific memory strategies.}
  \label{fig:copy}
\end{figure}

To gain insight into how the CMM leverages its STM and LTM, we visualize the Transformer's attention for various tasks, shown in~\cref{fig:copy}. On copy, STM queries attend predominantly to LTM. On sort, LTM attends to both stores during input, while STM increasingly attends to itself during output. On maze-solving tasks, STM queries attend entirely to the zero-valued sink, suppressing reads from both STM and LTM. This suggests that CMM's gains arise from the expressivity of the memory Transformers' feed-forward networks. On ImageNet-100, STM queries read from LTM but increasingly attend to the sink at later iterations. These patterns highlight the CMM's flexibility to draw on both stores or emphasize one according to the task. By augmenting the CTM's short-term temporal processing with a separately accessible persistent store, the CMM can leverage long-term memory when useful while retaining the ability to rely on short-term computation.

\section{Discussion and Conclusion}
\label{sec:conclusion}

The Continuous Memory Machine combines the expressive short-term processing of the CTM with the persistent long-term memory of MANNs. The CMM's two matrix-valued recurrent states are jointly processed by a Transformer, acting as an expressive bidirectional read--write mechanism. The CMM outperforms a suite of baselines, with improved performance in algorithmic, in-context learning and maze-solving tasks. The CMM improves upon the generalization capabilities of prior MANNs, while maintaining the interpretable attention patterns of the CTM. Ablations support the benefits of utilizing the CTM's short-term neuron-level processing, with the LSTM variant performing markedly worse throughout. Furthermore, analyzing the attention patterns of the CMM's memory Transformer highlights the flexibility of the CMM, with the model relying on LTM for algorithmic tasks and in-context learning, yet bypassing it when not required.

The benefits of the CMM come with additional costs and limitations. First, the CMM introduces additional hyperparameters, namely the LTM size $M_L$ and Transformer depth $J$. The joint memory update also increases computational and memory requirements, with training times and resource measurements reported in \cref{app:compute}. A further limitation, shared with nonlinear RNNs more broadly, is that recurrence restricts training parallelism compared with linear recurrent architectures such as the GDN. Recent methods for accelerating nonlinear RNN training therefore offer a promising direction for future work~\citep{danieli2026pararnn}. Beyond accelerating the recurrence itself, future work could also improve the efficiency of the CMM's memory update. In the current architecture, the memory Transformer is applied at every recurrent step, even when attention is routed primarily to the sink token and little cross-memory interaction is required. This suggests that updating the memory selectively, or only once every $t$ steps, could reduce the computational cost without requiring a memory update at every iteration.

More broadly, the CMM serves as an example of how principles drawn from biological memory can inspire neural architectures with capabilities beyond those exhibited by existing recurrent models, combining flexible short-term computation with persistent storage to improve both algorithmic generalization and recurrent reasoning.

\subsection*{AI use statement}
In this work we used generative AI tools for polishing the writing. Furthermore, we used AI as a coding tool for building out some of the baseline architectures based on the original papers and their corresponding codebases. One author has tested the correctness of the code. For example, when building out the GDN baseline~\citep{yang2024gated}, we verified that the introduction of negative eigenvalues~\citep{grazzi2025unlocking} allows the model to solve the parity task. We have not used AI for research ideation, the main writing of the text, performing the literature review, etc. We take responsibility for the final content of this work, including text, claims or artifacts produced with the aid of generative AI.

\subsection*{Reproducibility statement}
To ensure reproducibility of this work, we release all code at \url{https://github.com/SakanaAI/continuous-memory-machines}. The repository contains everything needed to reproduce the main results, including the baselines, for all tasks. Instructions on how to run each task are described in the \texttt{README.md} of the codebase. A single NVIDIA H100 was used for each experiment in the paper, and is sufficient to reproduce the results.

\bibliographystyle{unsrtnat}
\bibliography{main}

\appendix

\section{Architecture Details}
\label{app:architecture}

\subsection{CMM Details}

This section provides the full architectural specification of the CMM. For notational clarity in this section, we treat sequences as having tokens along the leading dimension, which is the transpose of the convention used in the main text.

\paragraph{Learnable Initialization.} The LTM is initialized with a learned matrix $B^0 \in \mathbb{R}^{M_L \times d}$, drawn from $\mathcal{N}(0, 0.01^2)$. Unlike the CTM, whose initial activated state and STM are both learned parameters, the CMM fixes both to zero buffers. The learned $B^0$ is the only trained initial state.

\paragraph{Sink Token.} A learned vector $s \in \mathbb{R}^{d}$, initialized to zero, is prepended to the concatenated sequence~(\cref{eq:cmm_concat}). Following~\citet{xiao2023efficient}, the sink absorbs uninformative attention mass, with its value projection zeroed at every block so that attention weight assigned to it contributes nothing to the output.

\paragraph{Positional Embeddings.} A learned matrix $P \in \mathbb{R}^{(1 + M_L + M_S) \times d}$, initialized from $\mathcal{N}(0, 0.02^2)$, encodes the slot identity (sink, LTM position, or STM position) of each token. The same $P$ is shared across all $J$ Transformer blocks.

\paragraph{Transformer Block.} Each of the $J$ blocks is a pre-norm residual block~\citep{xiong2020layer} with multi-head self-attention and a feedforward network (FFN). Given an input sequence $X \in \mathbb{R}^{S \times d}$ (where $S = 1 + M_L + M_S$) and positional embeddings $P$, block $j$ computes:
\begin{align}
    \bar{X} &= \mathrm{LayerNorm}(X), \\
    Q &= W_Q(\bar{X} + P), \quad K = W_K(\bar{X} + P), \quad V = W_V(\bar{X}), \label{eq:app_qkv} \\
    V_{0} &\leftarrow \mathbf{0}, \label{eq:app_sink_zero} \\
    X' &= X + \gamma_j\,\mathrm{Dropout}\!\Big(W_O \, \mathrm{Dropout}\!\Big(\mathrm{softmax}\!\Big(\frac{Q K^\top}{\sqrt{d_h}}\Big)\Big) V\Big), \label{eq:app_attn} \\
    X'' &= X' + \gamma_j\,\mathrm{Dropout}\!\Big(\mathrm{FFN}\!\big(\mathrm{LayerNorm}(X')\big)\Big), \label{eq:app_ffn}
\end{align}
where $W_Q, W_K, W_V, W_O \in \mathbb{R}^{d \times d}$ are bias-free linear projections split into $H$ heads, $d_h = d / H$ is the per-head dimension, and
\begin{equation}
    \mathrm{FFN}(x) = W_2\,\mathrm{SiLU}(W_1 x), \quad W_1, W_2 \in \mathbb{R}^{d \times d},
    \label{eq:app_ffn_def}
\end{equation}
also bias-free. Two design choices merit emphasis. First, the positional embeddings are added to the queries and keys but \emph{not} to the values~(\cref{eq:app_qkv}), so that position influences which tokens attend to one another but does not alter the content that is read. Second, the value vector of the sink token is set to zero at every block~(\cref{eq:app_sink_zero}), allowing the model to discard uninformative attention mass rather than distributing it across content-bearing tokens.

\paragraph{Residual Gate.} For Maze and Parity, each block has its own unconstrained learned scalar $\gamma_j$, initialized to zero and shared by its attention and FFN residual branches. Each block therefore starts as the identity map. For all other CMM experiments, $\gamma_j=1$ is fixed, giving ungated residual updates.

\paragraph{Output Processing.} After the final Transformer block, the output sequence is sliced as in~\cref{eq:cmm_transformer}. Layer normalization is applied to the updated LTM tokens before they are carried forward to the next timestep:
\begin{equation}
    \tilde{B}^t \leftarrow \mathrm{LayerNorm}(\tilde{B}^t).
    \label{eq:app_ltm_norm}
\end{equation}
The updated STM tokens are passed to the neuron-level models.

\paragraph{Neuron-Level Models.} The NLMs follow the same SuperLinear-GLU structure as the CTM~(\cref{eq:nlm}), with the addition of a LayerNorm applied to the activated state after the GLU~\citep{dauphin2017language}:
\begin{equation}
    z^{t+1} = \mathrm{LayerNorm}\!\Big(\mathrm{GLU}\!\big(W \cdot \tilde{A}^t\big)\Big),
    \label{eq:app_nlm}
\end{equation}
where $W \in \mathbb{R}^{M_S \times 2 \times d}$ is the per-neuron weight tensor. The SuperLinear layer applies an independent linear map to each neuron's $M_S$-length history, producing two outputs per neuron that are then halved by the GLU gate.

\subsection{Architectural Comparison}
\label{app:arch-comparison}

\begin{figure}[t!]
    \centering

    \begin{subfigure}{\linewidth}
        \centering
        \includegraphics[width=\linewidth]{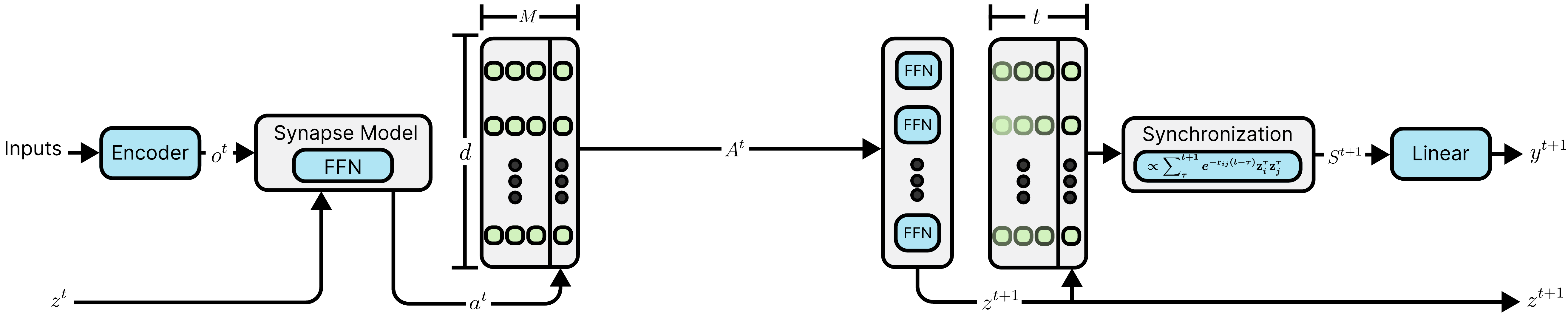}
        \caption{Continuous Thought Machine (CTM).}
        \label{fig:ctm_sideways_appendix}
    \end{subfigure}

    \vspace{1em}

    \begin{subfigure}{\linewidth}
        \centering
        \includegraphics[width=\linewidth]{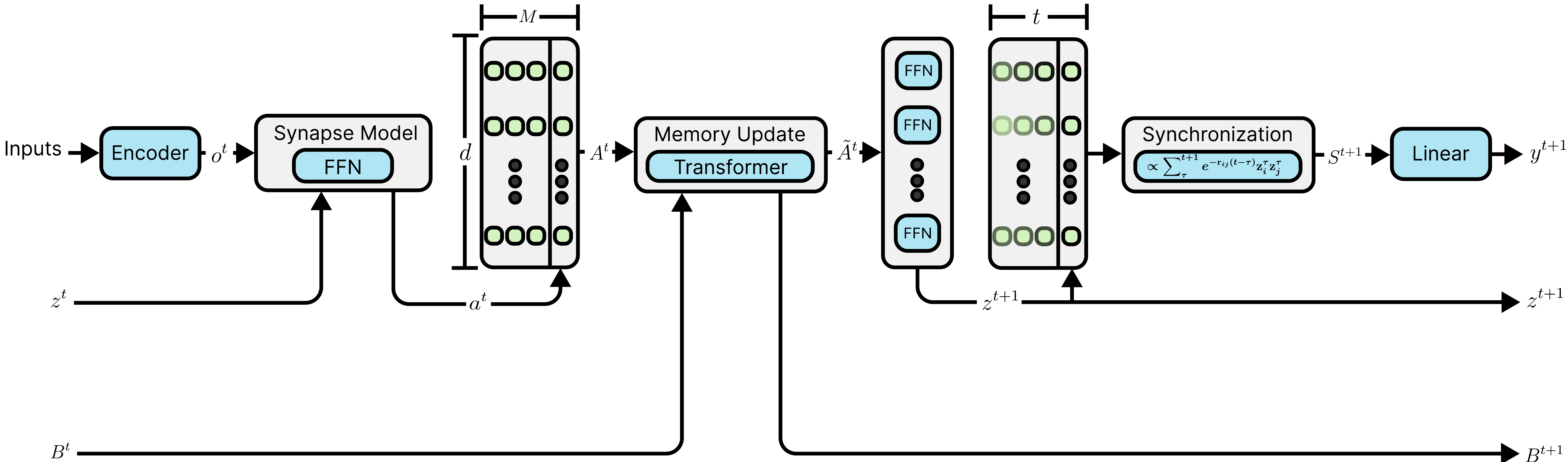}
        \caption{Continuous Memory Machine (CMM).}
        \label{fig:cmm_sideways_appendix}
    \end{subfigure}

    \vspace{1em}

    \begin{subfigure}{\linewidth}
        \centering
        \includegraphics[width=1\linewidth]{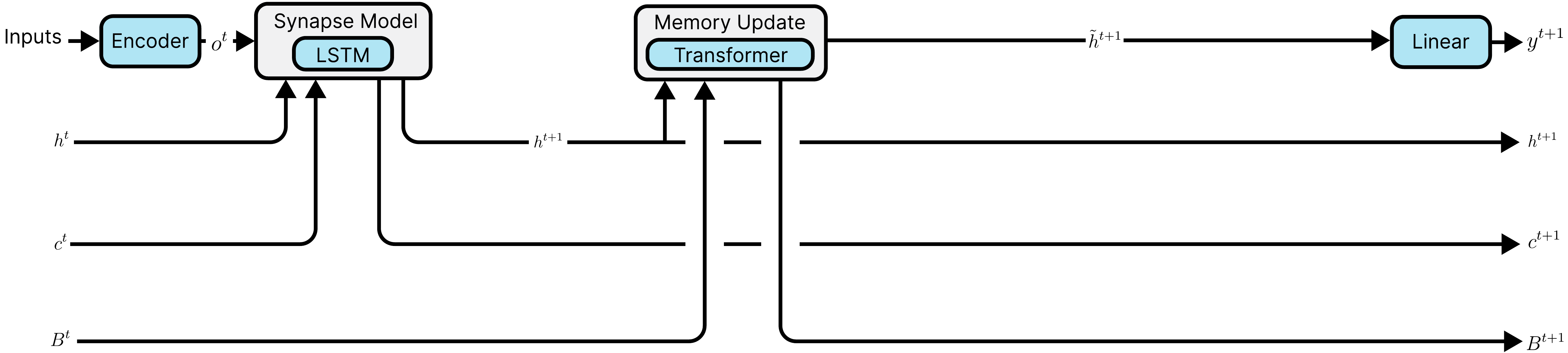}
        \caption{CMM-LSTM (LSTM substrate).}
        \label{fig:ab1_sideways_appendix}
    \end{subfigure}

    \caption{
    Comparison of the Continuous Thought Machine (top), the Continuous Memory Machine (middle), and CMM-LSTM (bottom). The CTM maintains a sliding short-term memory (STM) of recent pre-activations, which is processed by independent \textit{neuron-level} models, while the \textit{synchronization} module computes time-decayed outer products of activation histories that are projected to output logits. The CMM extends the CTM with a persistent long-term memory (LTM) that is jointly updated with the STM through the memory-update Transformer before the neuron-level models are applied. CMM-LSTM replaces the CTM backbone with a LayerNorm LSTM while retaining the same LTM and memory-update Transformer. Consequently, the Transformer receives a single hidden-state token instead of the CTM's sliding window of $M_S$ STM tokens, and there are no neuron-level models or synchronization representation. Comparing the CMM and CMM-LSTM isolates the contribution of the CTM's distributed short-term representation and per-neuron processing.
    }
    \label{fig:three-architectures}
\end{figure}

\begin{figure}[t!]
    \centering
    \begin{subfigure}[b]{0.45\textwidth}
        \centering
        \includegraphics[width=0.5\textwidth]{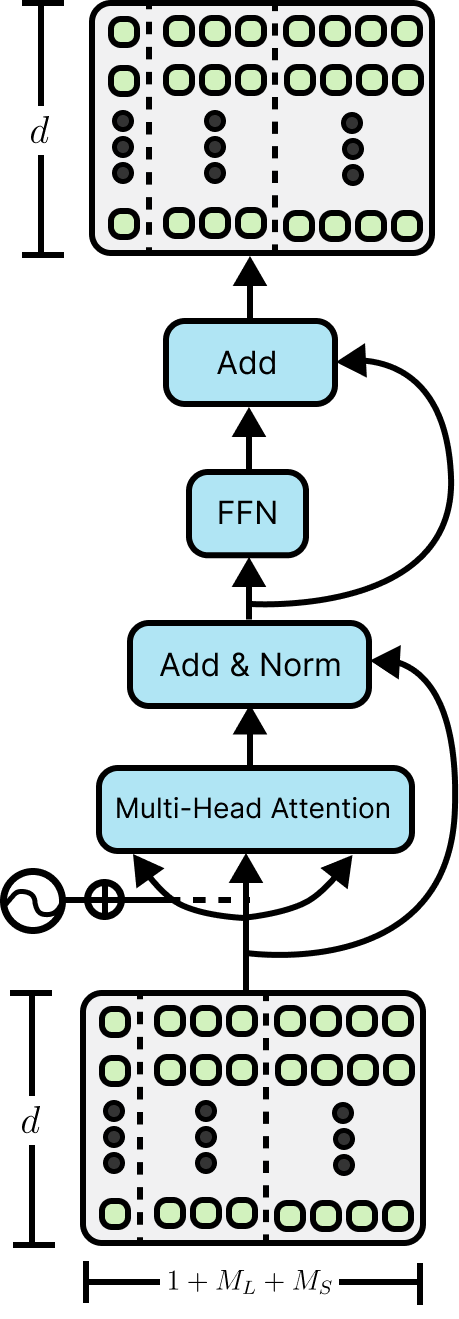}
        \caption{CMM memory update}
        \label{fig:cmm-transformer}
    \end{subfigure}
    \hfill
    \begin{subfigure}[b]{0.45\textwidth}
        \centering
        \includegraphics[width=0.5\textwidth]{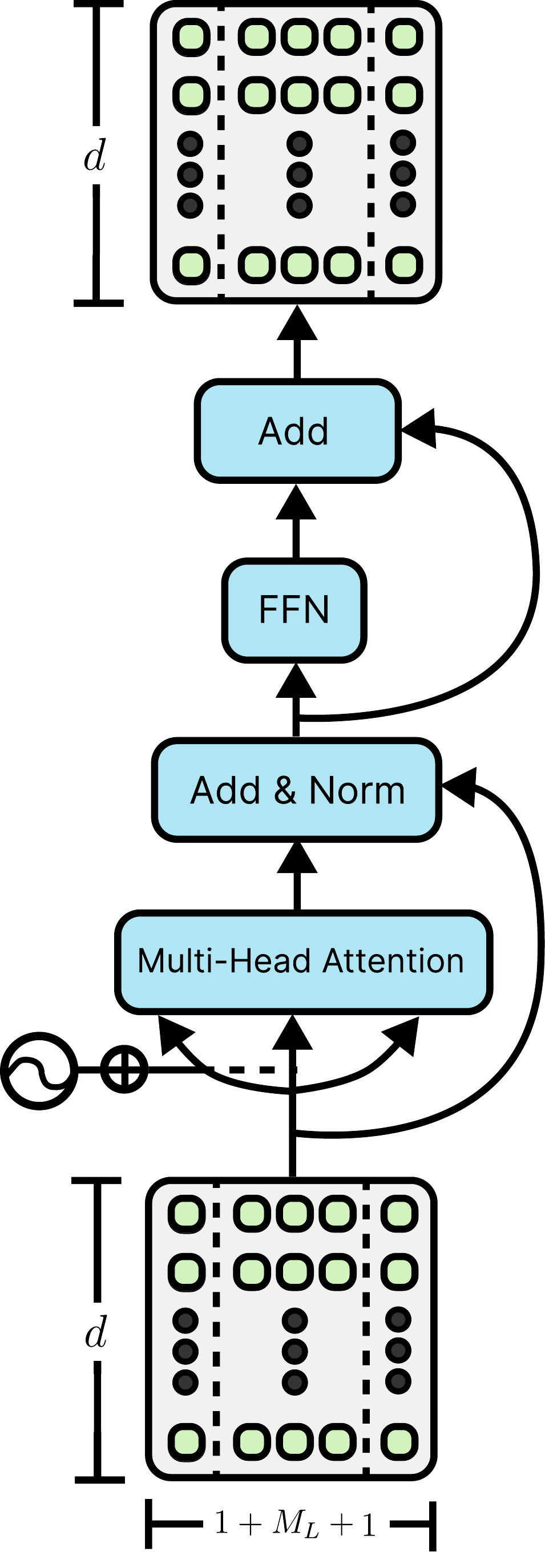}
        \caption{CMM-LSTM memory update}
        \label{fig:ab1-transformer}
    \end{subfigure}
    \caption{Memory-update modules for the CMM (a) and CMM-LSTM (b). Both apply a Transformer over a sequence consisting of a sink token, the LTM tokens, and STM tokens, with learned positional embeddings. The CMM (a) attends jointly over $M_S$ STM tokens drawn from the sliding window of recent pre-activations, giving the Transformer a multi-token view of recent activity. CMM-LSTM (b) replaces the sliding window with the LSTM's single hidden vector, so only one STM token is passed to the Transformer. Comparing the two isolates the effect of the CTM's distributed short-term representation on the memory update.}
    \label{fig:transformer-comparison}
\end{figure}

We provide a more detailed overview of the ablation described in~\cref{sec:baselines}. Side-by-side diagrams of the full architectures are shown in~\cref{fig:three-architectures}, and a closer comparison of the memory-update modules is shown in~\cref{fig:transformer-comparison}. The CMM and CMM-LSTM share an identical LTM and memory-update Transformer; they differ only in how the short-term state is represented. The CMM passes the full sliding window of $M_S$ recent pre-activations to the Transformer, whereas CMM-LSTM passes the LSTM's single compressed hidden vector. Any performance gap between the two therefore reflects the contribution of the CTM's distributed short-term representation and per-neuron processing, rather than the LTM or Transformer alone.

\subsection{Synchronization Normalization}
\label{app:synch_norm}
The original CTM~\citep{darlow2025continuous} normalizes the synchronization matrix by dividing by the square root of the sum of exponential weights:
\begin{equation}
    S^{t+1}_{ij} \;=\;
    \frac{\textstyle\sum_{\tau=1}^{t+1} \mathrm{e}^{-r_{ij}(t-\tau)}\, z^{\tau}_i\, z^{\tau}_j}
         {\left(\textstyle\sum_{\tau=1}^{t+1}\mathrm{e}^{-r_{ij}(t-\tau)}\right)^{1/2}}.
\end{equation}
In our implementation, we remove the square root from the denominator. The square root causes the numerator and denominator to grow at different rates with $t$, so the magnitude of $S^t_{ij}$ becomes implicitly tied to the number of internal ticks. Dividing by the linear sum of weights instead yields a properly normalized exponential moving average whose expected magnitude is invariant to $t$, allowing the model to generalize to thinking horizons beyond those seen during training.

We compute synchronization for $d$ neuron pairs, sampled uniformly without replacement from the $d^2$ ordered pairs at initialization and then held fixed. After dividing by the linear sum of weights, we add $10^{-8}$ to each component and apply RMSNorm~\citep{zhang2019root} with learned elementwise scaling to obtain the $d$-dimensional synchronization representation.

\section{Additional Ablations}
\label{app:additional-ablations}

\subsection{Memory capacity and update depth}
\label{app:sweeps}
\begin{figure}[htbp]
    \centering
    \begin{subfigure}[b]{0.48\linewidth}
        \centering
        \includegraphics[width=\linewidth]{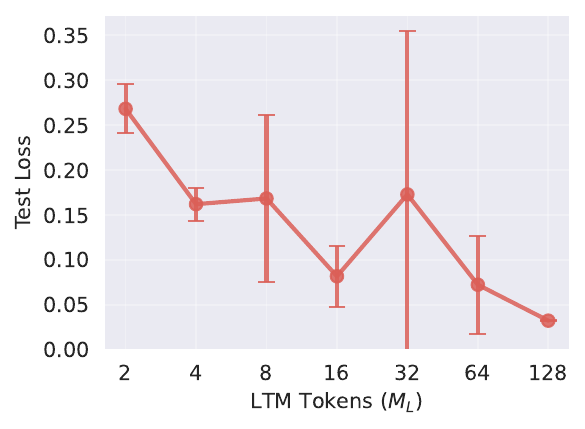}
        \caption{LTM token sweep.}
        \label{fig:priority-sort-ltm-sweep}
    \end{subfigure}
    \hfill
    \begin{subfigure}[b]{0.48\linewidth}
        \centering
        \includegraphics[width=\linewidth]{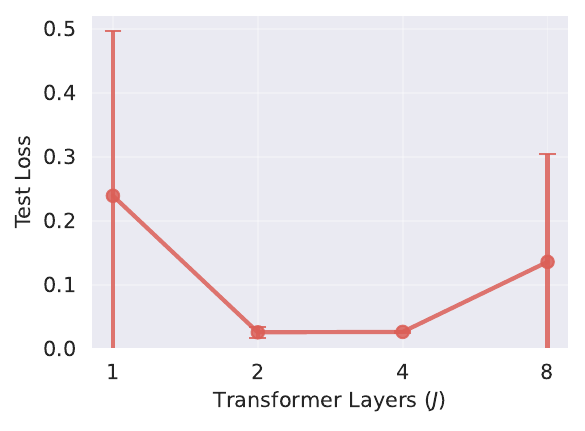}
        \caption{Transformer depth sweep.}
        \label{fig:priority-sort-depth-sweep}
    \end{subfigure}
    \caption{Additional CMM ablations on Priority Sort. Points show mean test binary cross-entropy. Error bars show one standard deviation over two seeds. Lower is better.}
    \label{fig:priority-sort-additional-ablations}
\end{figure}

We evaluate how Priority Sort performance changes with long-term memory capacity and memory update depth. For the capacity sweep, we reduce the STM length to $M_S{=}5$ to limit short-term storage, fix the memory update Transformer depth at $J{=}2$, and vary $M_L\in\{2,4,8,16,32,64,128\}$. For the depth sweep, we use the standard Priority Sort configuration with $M_S{=}20$ and $M_L{=}128$, and vary $J\in\{1,2,4,8\}$. All runs use the same 500K parameter budget and training setup as the main Priority Sort experiment. As shown in~\cref{fig:priority-sort-additional-ablations}, increasing LTM capacity broadly improves performance when STM is restricted, while two to four Transformer layers perform best in the depth sweep.

\section{Compute Resources}
\label{app:compute}
Experiments were run on a single NVIDIA H100 GPU. Mean training-loop wall times per CMM seed are listed in \cref{table:compute}. These include periodic evaluation and checkpointing, and exclude model setup, queueing, and downtime between resumed jobs. For Maze, we sum the original 4,500-epoch run and its 4,500-epoch continuation; both use 50 internal iterations, giving 9,000 epochs in total. Means are taken over three seeds, so total GPU-hours are three times the per-seed values, even when seeds run concurrently. Baseline architectures were trained under matched conditions. The full research project required additional compute beyond the reported experiments, including preliminary architecture exploration, hyperparameter tuning, and ablations that are not included in the final paper.

To characterize computational cost, we benchmark all architectures on batches of 64 complete Copy-100 sequences, each containing 100 input and 100 output steps. Hidden widths are selected independently under a 500K parameter budget, including a linear $12\!\to\!64$ input projection and a 10-dimensional output projection. Measurements use eager float32 execution on a single NVIDIA H100 GPU, with five warmup repetitions followed by twenty measured repetitions per timing. Each architecture runs in a fresh process. Forward timing disables autograd; forward+backward timing differentiates the sum of all outputs and excludes optimizer updates. \Cref{table:timing-memory} reports the mean latencies and memory use. The CMM's joint STM--LTM update increases computation and activation memory relative to the RNN and LSTM.

\begin{table}[h]
  \caption{Mean training-loop wall time per CMM seed on a single NVIDIA H100 GPU.}
  \label{table:compute}
  \centering
  \begin{tabular}{lrc}
    \toprule
    Experiment & Epochs & Time per seed \\
    \midrule
    Copy & 1000 & 6.9 h \\
    Associative Recall & 1000 & 1.1 h \\
    Sort & 1000 & 3.0 h \\
    Sort (length gen.) & 3000 & 2.2 h \\
    Parity & 200 & 0.6 h \\
    Few-Shot Regression & 1000 & 1.3--1.8 h \\
    ImageNet-100 & 25 & 0.9 h \\
    Maze & 9000 & 127.5 h \\
    \bottomrule
  \end{tabular}
\end{table}

\begin{table}[h]
  \caption{GPU efficiency on complete Copy-100 sequences (batch size 64, sequence length 200). Times are amortized milliseconds per step, obtained by dividing full-sequence latency by 200; the Transformer processes the sequence in parallel and RMT uses ten-step segments. State is persistent memory per example (KiB); peak is allocated GPU memory for a complete forward+backward pass (MiB). The sequence Transformer has no recurrent state.}
  \label{table:timing-memory}
  \centering
  \begin{small}
  \setlength{\tabcolsep}{3pt}
  \begin{tabular}{lrrrrrr}
    \toprule
    Architecture & Params & $d$ & Fwd/step & Fwd+Bwd/step & State & Peak \\
    \midrule
    \textcolor{black!60}{Transformer} & \textcolor{black!60}{479{,}278} & \textcolor{black!60}{100} & \textcolor{black!60}{0.011} & \textcolor{black!60}{0.032} & \textcolor{black!60}{--} & \textcolor{black!60}{639.4} \\
    \midrule
    RNN & 499{,}170 & 668 & 0.066 & 0.246 & 2.61 & 201.4 \\
    LSTM & 497{,}558 & 318 & 0.147 & 0.548 & 2.48 & 336.2 \\
    FS-LSTM & 494{,}926 & 149 & 0.450 & 1.473 & 2.33 & 415.4 \\
    GDN & 495{,}946 & 480 & 0.555 & 2.126 & 275.62 & 7{,}248.0 \\
    DNC & 498{,}220 & 271 & 1.266 & 5.097 & 78.20 & 3{,}175.0 \\
    RMC & 479{,}822 & 108 & 0.412 & 1.464 & 13.50 & 3{,}079.1 \\
    LSAM & 496{,}122 & 372 & 0.219 & 0.976 & 136.59 & 3{,}644.1 \\
    NAM-TM & 496{,}992 & 170 & 0.960 & 4.706 & 67.33 & 2{,}039.4 \\
    RMT & 496{,}682 & 160 & 0.255 & 0.856 & 80.00 & 8{,}646.6 \\
    CTM & 498{,}802 & 472 & 0.360 & 1.256 & 25.81 & 634.5 \\
    \midrule
    \textcolor{black!60}{Ab: CMM-LSTM} & \textcolor{black!60}{491{,}054} & \textcolor{black!60}{204} & \textcolor{black!60}{0.529} & \textcolor{black!60}{1.793} & \textcolor{black!60}{27.09} & \textcolor{black!60}{4{,}587.7} \\
    \textcolor{black!60}{Ab: CTM-Ext.} & \textcolor{black!60}{498{,}092} & \textcolor{black!60}{390} & \textcolor{black!60}{0.359} & \textcolor{black!60}{1.257} & \textcolor{black!60}{216.33} & \textcolor{black!60}{3{,}145.8} \\
    \textcolor{black!60}{Ab: CMM-STM} & \textcolor{black!60}{495{,}862} & \textcolor{black!60}{212} & \textcolor{black!60}{0.850} & \textcolor{black!60}{2.755} & \textcolor{black!60}{117.59} & \textcolor{black!60}{21{,}325.3} \\
    CMM (Ours) & 486{,}474 & 224 & 0.870 & 2.813 & 124.25 & 22{,}330.1 \\
    \bottomrule
  \end{tabular}
  \end{small}
\end{table}

\section{Experimental Details}
\label{app:experimental-details}

Unless otherwise noted in a task-specific subsection, experiments use the AdamW optimizer~\citep{kingma2014adam} with a learning rate of $10^{-4}$ and no learning rate scheduler. ImageNet-100 uses cosine annealing over its 25 training epochs. Weight decay~\citep{krogh1991simple} is set to 0 except for ImageNet-100 and Priority Sort length generalization, where it is $10^{-4}$. Models use parameter budgets of 500K by default, 1M for ImageNet-100 (excluding the frozen backbone), 10M for Maze, and 60K for Parity and Priority Sort length generalization. For Maze and Parity, each CMM memory update Transformer block uses a learned scalar initialized to zero. The scalar scales both the attention and feed-forward residual branches. All other CMM experiments use ungated residual updates. \Cref{table:shared-hyperparams} summarizes the remaining training hyperparameters for each task. \Cref{table:cmm-hyperparams} lists the CMM-specific architectural hyperparameters used for each task. Task-specific details follow in the subsequent sections. The error bars in \cref{fig:algorithmic-tasks-a,fig:algorithmic-tasks-b} show one standard deviation over three seeds.

\begin{table}[h]
  \caption{Per-task training hyperparameters.}
  \label{table:shared-hyperparams}
  \centering
  \begin{small}
  \begin{tabular}{lcccc}
    \toprule
    Task & Loss & Num. Epochs & Batch Size & Grad. Clip \\
    \midrule
    Copy & MSE & 1000 & 128 & 5 \\
    Associative Recall & BCE & 1000 & 32 & 50 \\
    Priority Sort & BCE & 1000 & 32 & 50 \\
    Parity & BCE & 200 & 64 & 5 \\
    Few-Shot Regression & MSE & 1000 & 64 & 5 \\
    ImageNet-100 & CE & 25 & 64 & 1 \\
    Maze & Dual-tick CE\textsuperscript{$\dagger$} & 9000 & 1024 & 1 \\
    \bottomrule
  \end{tabular}
  \end{small}
  \vspace{2pt}

  {\footnotesize \textsuperscript{$\dagger$}Dual-tick selection with auto-curriculum masking over 5 action classes (see~\cref{app:maze}).}
\end{table}

\begin{table}[h]
  \caption{CMM-specific hyperparameters for each task. $M_S$: STM length (sliding window), $M_L$: number of LTM tokens, $J$: Transformer layers in the memory update, $D$: synapse depth.}
  \label{table:cmm-hyperparams}
  \centering
  \begin{small}
  \begin{tabular}{lcccc}
    \toprule
    Task & $M_S$ & $M_L$ & $J$ & $D$ \\
    \midrule
    Copy & 10 & 128 & 1 & 8 \\
    Associative Recall & 10 & 32 & 1 & 8 \\
    Priority Sort & 20 & 128 & 2 & 8 \\
    Parity & 4 & 4 & 1 & 8 \\
    Few-Shot Regression & 10 & 32 & 1 & 8 \\
    ImageNet-100 & 10 & 32 & 1 & 4 \\
    Maze & 10 & 8 & 1 & 16 \\
    \bottomrule
  \end{tabular}
  \end{small}
\end{table}

\paragraph{Transformer baseline.}
On Copy, Associative Recall, and Priority Sort, the Transformer uses four independently parameterized causal pre-normalization layers with hidden dimension 100, four attention heads, GELU feed-forward sublayers of dimension 384, fixed sinusoidal positional embeddings, a final layer normalization, and no dropout. The maximum context lengths are 256 for Copy, 32 for Associative Recall, and 64 for Priority Sort, in each case covering the complete input--output sequence. The respective models have 479{,}278, 478{,}554, and 479{,}012 trainable parameters and otherwise use the shared task-specific optimization settings in~\cref{table:shared-hyperparams}. The smaller Transformer used for length generalization is specified in~\cref{app:priority-sort}.

\paragraph{Recurrent Memory Transformer baseline.}
The RMT~\citep{bulatov2022recurrent} uses two layers, four attention heads, relative positional attention, post-layer normalization, and ReLU feed-forward sublayers of width $2d$. Memory tokens pass information between segments. We use the shared task-specific optimization settings and report final-epoch results over three seeds. \Cref{table:rmt-hyperparams} lists the resolved configurations. On ImageNet-100, each segment corresponds to one visual iteration, and accuracy is pooled over all 5,000 test images per seed.

\begin{table}[htbp]
  \caption{RMT configurations. Parameter counts exclude the frozen ImageNet backbone.}
  \label{table:rmt-hyperparams}
  \centering
  \begin{small}
  \begin{tabular}{lrrrr}
    \toprule
    Task & $d$ & Segment length & Memory tokens & Parameters \\
    \midrule
    Copy & 160 & 10 & 128 & 496,682 \\
    Associative Recall & 160 & 10 & 32 & 480,358 \\
    Priority Sort & 160 & 20 & 128 & 496,296 \\
    Priority Sort (length gen.) & 52 & 4 & 22 & 54,752 \\
    Parity & 56 & 4 & 4 & 59,569 \\
    Few-Shot Regression & 160 & 10 & 32 & 496,801 \\
    ImageNet-100 & 152 & 1 & 32 & 997,876 \\
    \bottomrule
  \end{tabular}
  \end{small}
\end{table}

\subsection{Copy Task}
\label{app:copy}

The model receives a sequence of $T{=}100$ vectors of dimension $D{=}10$ with entries drawn uniformly from $[-1, 1]$ and must reproduce the sequence immediately after the input phase. The input at each step is $(r_t, s_t, q_t)$ where $r_t$ is the data vector (zero-padded outside the input phase), $s_t$ is a binary sample flag active during the input phase, and $q_t$ is a binary query flag active during the output phase. The total sequence length is $2T$: $T$ input steps followed by $T$ query steps. A linear projection maps the $(D{+}2)$-dimensional input to the model's hidden dimension. The loss is mean squared error computed only over the query phase. The dataset consists of 6,400 procedurally generated episodes per epoch.

\subsection{Associative Recall}
\label{app:assoc-recall}

Each trial presents $N$ items, with $N$ sampled uniformly from $\{2,\ldots,6\}$, each consisting of 3 binary vectors of 6 bits. The item sequence is zero-padded to 18 vector steps, followed by a delimiter, a three-vector query item, a second delimiter, and three output slots. The query is sampled uniformly from the first $N-1$ items, and the model must output its successor in the original sequence. Binary cross-entropy is computed only over the three output vectors. The dataset consists of 6,400 procedurally generated episodes per epoch.

\subsection{Priority Sort}
\label{app:priority-sort}

The model receives $N{=}20$ input items, each consisting of a scalar priority drawn uniformly from $[-1,1]$ and an associated 8-bit binary value vector. After the input phase, the model must output the 16 value vectors corresponding to the 16 lowest-priority items, sorted in ascending order of priority. The loss is binary cross-entropy computed over the output positions. The dataset consists of 6,400 procedurally generated episodes per epoch.

\paragraph{Length generalization.} For \cref{fig:algorithmic-tasks-b}, we use $W=4$, $K=4$, and three training seeds per model. Each training batch contains a single input length, with lengths $N\in\{4,\ldots,16\}$ balanced across the 13 batches of 64 episodes in each epoch. All models train for 3,000 epochs under a 60K parameter budget, using AdamW with learning rate $10^{-3}$, weight decay $10^{-4}$, dropout $0.1$, and gradient clipping at 50. We evaluate the final checkpoints on 832 freshly generated episodes at each $N\in\{4,8,12,16,20,30,40,50\}$, using the same evaluation examples across models and training seeds. Bit accuracy is computed over the four output value vectors; unbiased random bit predictions have expected accuracy $50\%$. The Transformer has two causal pre-normalization layers, hidden dimension 56, four attention heads, GELU feed-forward sublayers of dimension 128, fixed sinusoidal positional embeddings with a maximum sequence length of 2,048, and a final layer normalization. It has 59{,}068 trainable parameters. The RMT uses two layers with four attention heads, hidden dimension 52, ReLU feed-forward sublayers of width 104, four input steps per segment, and 22 recurrent memory tokens, for 54{,}752 trainable parameters.

\subsection{Parity}
\label{app:parity}

Each example contains independent uniformly sampled bits followed by a query token. The target is one if the number of set bits is odd and zero otherwise. Binary cross-entropy is computed only at the query step. All models use three seeds, a 60K parameter budget, zero dropout, and 200 training epochs with 8,192 generated examples per epoch. Training lengths vary from 1 to 40. We evaluate on 512 examples at each length in $\{40,50,100,200,500,1000\}$ and report the final epoch. The CMM uses $M_S=4$ and $M_L=4$; the standard CTM uses $M_S=4$, and CTM-Ext.\ uses $M_S=8$ to match the CMM's total memory capacity.

\subsection{Few-Shot Regression}
\label{app:fewshot-regression}

In each episode a new target linear function $f: [-1,1]^{n} \to \mathbb{R}$ with $n{=}12$ is sampled. The model receives $K$ support pairs $(x_t, f(x_t) + \epsilon_t)$ with $\epsilon_t \sim \mathcal{N}(0, 0.01)$ sequentially, followed by 20 query inputs for which it must predict outputs. We evaluate with $K \in \{10, 20\}$. Training minimizes MSE over the complete episode. Evaluation reports accuracy only on the 20 query predictions: the fraction satisfying $|\hat y_t-y_t|<0.05$, pooled over all test episodes. The clean targets $y_t$ are standardized to zero mean and unit sample standard deviation within each episode before noise is added to the support-value inputs. Thus the threshold is $0.05$ target standard deviations; support predictions do not enter the reported accuracy. At each step, the 12 input variables, observed-value channel, and support-mask bit are mapped to the model's hidden dimension by a learned linear projection followed by a ReLU activation. The dataset consists of 10,000 episodes, with fresh functions sampled each epoch.

CTM-Ext.\ uses $M_S{=}42$, matching the CMM's $10$ STM slots plus $32$ LTM tokens. CMM-STM uses the same $42$-slot STM with one four-head memory-Transformer layer and no persistent LTM. Hidden width is selected independently under the shared 500K parameter budget, yielding 499{,}585 parameters for CTM-Ext.\ and 495{,}713 for CMM-STM, compared with 496{,}129 for the CMM. Both ablations use the shared optimization settings in~\cref{table:shared-hyperparams}. We report the mean and one sample standard deviation of final epoch-1{,}000 test query accuracy across seeds 0, 1, and 2. Each seed is evaluated on 10,000 test episodes, giving 200,000 query predictions.

The Transformer baseline uses four independently parameterized causal pre-normalization layers with hidden dimension 104, four attention heads, and GELU feed-forward sublayers of dimension 384. It uses fixed sinusoidal positional embeddings, a maximum sequence length of 64, a final layer normalization, and no dropout. The causal mask restricts each position to the current and preceding items. A linear head maps each output representation to a scalar prediction. The model has 498{,}033 trainable parameters and otherwise uses the shared optimization settings in~\cref{table:shared-hyperparams}.

\subsection{Image Classification (ImageNet-100)}
\label{app:imagenet}

We use the ImageNet-100 subset containing 100 classes with images resized to $224 \times 224$. We extract $14 \times 14 = 196$ patch tokens of dimension 384 from the output of the penultimate Transformer block (block 11 of 12) of a frozen, pre-trained ViT-Small~\citep{wang2026vit}. These patches serve as keys and values for cross-attention from synchronization-based queries, as in the original CTM~\citep{darlow2025continuous}. The recurrent model processes the image over $I{=}25$ internal iterations, producing a class prediction at each step. The loss is standard cross-entropy applied to the final prediction. The trainable model uses a 1M parameter budget, excluding the frozen backbone. We train for 25 epochs with cosine annealing from an initial learning rate of $10^{-4}$ to zero, dropout of 0.2, and weight decay of $10^{-4}$. CTM-Ext.\ uses $M_S=42$, matching the CMM's 10 STM slots plus 32 LTM tokens, under the same 1M parameter budget. We report final-epoch accuracy pooled over all 5,000 test images per seed, with means and sample standard deviations over seeds 0, 1, and 2.

\subsection{Maze-Solving}
\label{app:maze}

We evaluate on the large maze dataset ($39{\times}39$ grids)~\citep{darlow2025continuous}. Each maze is encoded as a 3-channel image (walls, start, goal) and processed by a ResNet-34 backbone (first 2 blocks, no pre-training) producing $10 \times 10 = 100$ spatial tokens. The model recurrently processes these tokens over $I{=}50$ internal iterations, outputting at each step a route of length 100 as a sequence of 5-class action predictions (up, down, left, right, wait).

The loss uses dual-tick selection: for each sample we compute the per-route-step cross-entropy at every internal tick, then average the loss at the minimum-CE tick and the most-certain tick (the tick with lowest mean output entropy over route steps). An auto-curriculum mask restricts the loss to the first $c$ route steps beyond the model's current longest fully-correct prefix, where $c{=}25$. This encourages the model to learn incrementally longer routes.

We target 9,000 epochs with an effective batch size of 1,024 and learning rate $10^{-3}$. NAM-TM uses microbatches of 256 with four gradient accumulation steps; the other reported architectures use batches of 1,024 without accumulation. All models use dropout 0.1. The CMM memory update Transformer uses one attention head. \Cref{table:maze-results} reports the arithmetic mean and sample standard deviation across seeds 0, 1, and 2 of the final available current-model test accuracies, without EMA or selection of the best epoch during training. Predictions use each example's most-certain internal iteration. Action accuracy measures the fraction of correct route actions; board (full-maze) accuracy requires all 100 actions, including wait padding, to match the target. The recorded test metrics average batch accuracies over the 5,000 test mazes.

\Cref{table:maze-results-per-seed} lists both test accuracies for each seed. NAM-TM and CMM-STM completed all three seeds at epoch 9,000. RMC seed 2 stopped after a gradient spike during epoch 5,076; we retain its last recorded test evaluation at epoch 5,075. Every other run included in the tables ends at epoch 9,000.

\paragraph{Incomplete baselines.} RMT likewise exhibited unstable training and did not complete the 9,000-epoch schedule, so we omit its Maze results. For LSAM and GDN, the reference per-device batch size of 1,024 exceeded the available GPU memory. Smaller microbatches reduced memory usage, but matching the effective batch size required multiple sequential forward and backward passes per optimizer update. The resulting runtime made the full 9,000-epoch schedule impractical within our compute budget. We therefore omit LSAM and GDN from the Maze tables.

\begin{table}[htbp]
    \centering
    \caption{Final available test action and board (full-maze) accuracies (\%) for each training seed. Ending epochs are listed in the text; aggregate results appear in \cref{table:maze-results}.}
    \label{table:maze-results-per-seed}
    \small
    \setlength{\tabcolsep}{6pt}
    \begin{tabular}{@{}lrrrrrr@{}}
        \toprule
        & \multicolumn{2}{c}{Seed 0} & \multicolumn{2}{c}{Seed 1} & \multicolumn{2}{c}{Seed 2} \\
        \cmidrule(lr){2-3}\cmidrule(lr){4-5}\cmidrule(l){6-7}
        Architecture & Action & Board & Action & Board & Action & Board \\
        \midrule
        \textcolor{black!60}{Transformer} & \textcolor{black!60}{$33.82$} & \textcolor{black!60}{$0.20$} & \textcolor{black!60}{$43.40$} & \textcolor{black!60}{$1.10$} & \textcolor{black!60}{$47.09$} & \textcolor{black!60}{$0.84$} \\
        \midrule
        RNN & $49.63$ & $2.27$ & $46.67$ & $2.34$ & $45.25$ & $1.80$ \\
        LSTM & $39.67$ & $1.06$ & $52.66$ & $2.51$ & $59.46$ & $4.47$ \\
        FS-LSTM & $59.53$ & $4.10$ & $61.23$ & $4.58$ & $43.48$ & $1.08$ \\
        DNC & $81.44$ & $12.75$ & $81.00$ & $12.04$ & $85.76$ & $16.70$ \\
        RMC & $25.26$ & $0.00$ & $25.10$ & $0.00$ & $20.59$ & $0.00$ \\
        NAM-TM & $83.13$ & $13.39$ & $82.62$ & $13.64$ & $81.70$ & $12.68$ \\
        CTM & $86.76$ & $18.20$ & $97.93$ & $81.95$ & $81.64$ & $11.96$ \\
        \midrule
        \textcolor{black!60}{Ab: CMM-LSTM} & \textcolor{black!60}{$46.64$} & \textcolor{black!60}{$1.57$} & \textcolor{black!60}{$54.04$} & \textcolor{black!60}{$3.52$} & \textcolor{black!60}{$38.27$} & \textcolor{black!60}{$1.24$} \\
        \textcolor{black!60}{Ab: CMM-STM} & \textcolor{black!60}{$98.32$} & \textcolor{black!60}{$86.64$} & \textcolor{black!60}{$95.82$} & \textcolor{black!60}{$42.92$} & \textcolor{black!60}{$66.75$} & \textcolor{black!60}{$1.74$} \\
        CMM (Ours) & $97.90$ & $71.28$ & $91.93$ & $22.84$ & $99.08$ & $90.73$ \\
        \bottomrule
    \end{tabular}
\end{table}

\end{document}